\documentclass{article}

\usepackage[accepted]{icml2026}

\usepackage{amsmath}
\usepackage{amssymb}
\usepackage{amsfonts}
\usepackage{mathtools}
\usepackage{amsthm}
\usepackage{nicefrac}

\usepackage{graphicx}
\usepackage{subcaption}
\usepackage{wrapfig}
\usepackage{rotating}
\usepackage{capt-of}
\usepackage{float}
\usepackage{placeins}

\usepackage{hyperref}
\usepackage{booktabs}
\usepackage{tabularx}
\usepackage{longtable}
\usepackage{multirow}
\usepackage{threeparttable}

\usepackage[utf8]{inputenc}
\usepackage[T1]{fontenc}
\usepackage{microtype}
\usepackage{url}
\usepackage{xcolor}
\usepackage[normalem]{ulem}

\usepackage{ragged2e}
\newcolumntype{Y}{>{\Centering\arraybackslash}X}

\usepackage{spverbatim}
\usepackage[textsize=tiny]{todonotes}
\usepackage[frozencache,cachedir=.]{minted}

\usepackage[capitalize,noabbrev]{cleveref}

\theoremstyle{plain}

\theoremstyle{definition}

\theoremstyle{remark}

\hypersetup{
  pdftitle={Evaluating Language Models in Realistic Conversational Contexts},
  pdfauthor={Ilija Subasic, Andrew Rabinovich, Zhao Chen}
}

\begin{document}
\twocolumn[

  \icmltitle{Evaluating Language Models in Realistic Conversational Contexts}




  \begin{icmlauthorlist}
    \icmlauthor{Ilija Subasic}{yyy}
    \icmlauthor{Andrew Rabinovich}{yyy}
     \icmlauthor{Zhao Chen}{yyy}

  \end{icmlauthorlist}

  \icmlaffiliation{yyy}{Upwork Inc., Palo Alto, CA 94301, USA }

  \icmlcorrespondingauthor{Ilija Subasic}{ilija.subasic@cloud.upwork.com}

  \vskip 0.3in
]
\printAffiliationsAndNotice

%

\begin{abstract}

As Large Language Models (LLMs) are increasingly deployed to serve open-ended, multi-turn interactions, evaluating conversational quality at human scale has become a central challenge. Existing evaluation frameworks built for summarization, translation, or short-form QA tasks fall short of adequately measuring the consistency of human-scale dialogue, especially when derivation and validation of these metrics themselves often rely on synthetic rather than human sources. We fill the gap by introducing \textbf{UPHELD} (UPwork Human-Scale Evaluated Long Dialogues), a large, reference-full benchmark for evaluating human-scale conversational ability beyond factual correctness. UPHELD consists of hundreds of complete human-to-human dialogues authored by \textbf{professional script writers}, with realistic turn densities and \textbf{36,000+ per-turn human annotations} across \textbf{30,000+ expert-generated dialogue turns}. Using UPHELD, we systematically evaluate classical automatic metrics and reference-free LLM-as-a-judge approaches, and find them unreliable when correlated with expert human judgment. Building off this analysis, we use UPHELD to develop a \textbf{Mixture-of-Judges} framework that combines multiple evaluative signals and improves correlation with human assessments by approximately \textbf{30\%}. Overall, UPHELD provides a robust, human-grounded foundation for evaluating human-scale conversational intelligence that fills a crucial gap in the pre-existing LLM dataset landscape.

\end{abstract}

\section{Introduction}

The rapid advancement of Large Language Models (LLMs) has revolutionized text generation, particularly in focused, single-turn tasks like question answering. However, as these models are increasingly integrated into applications serving complex systems, the frontier of evaluation has shifted toward open-ended, multi-turn settings. In these environments, conversations are longer, less constrained, and more open-ended, and success is not merely defined by factual correctness but by the model's ability to maintain a natural, coherent, and useful interaction with a human end-user over an extended dialogue.
Existing evaluation metrics, largely inherited from pre-LLM natural language tasks such as machine translation, summarization, and question answering, focus on factual precision and lexical overlap. These metrics generally prove inadequate when assessing the nuanced qualities of a successful conversation, which depends on softer, long-range qualities like coherence, engagement, and the ability to maintain consistent tone over multiple turns.
While recent benchmarks have evaluated longer dialogues through reference-free ``LLM-as-a-judge'' evaluations \cite{zheng2023judgingllmasajudgemtbenchchatbot, g-eval, Dubois2023AlpacaFarmAS} or human preference platforms \cite{chiang2024chatbot}, these methods face critical limitations. LLM judge quality often fails to exceed that of non-expert human judges \cite{Bavaresco2024} \cite{no-free-ref}, and such benchmarks are increasingly susceptible to training data leakage \cite{Mirzadeh2024} and poor reliability on out-of-distribution tasks \cite{no-free-ref}. 
Additionally, many of the metrics are tested primarily on synthetic datasets, which can introduce significant biases in metric results due to empirically verified limits in LLM dialogue generation \cite{Wang2025Limits} and persistent biases in LLM-simulated data used for both training and evaluation \cite{Rahmani2025Biases}.

The work presented here mitigates these issues through a specialized benchmark evaluating models on conversations, which are designed to be:
\begin{itemize}
    \item \textbf{Multi-turn and human-to-human.} Emphasizing naturalistic flow of real interaction rather than simulated or distilled outputs.
    \item \textbf{Not anchored to specific external knowledge.} Shifting the focus from retrieval-based accuracy to conversational intelligence.
    \item \textbf{Not anchored to a specific outcome.} Allowing for fluid, open-ended dialogue without focusing on a ``right answer" (e.g. coding, summarization), but still having an overall goal (e.g., ``tell me how to make a pizza'').
\end{itemize}

We introduce \textbf{UPHELD (UPwork Human-Scale  Evaluated Long Dialogues)}, a novel collection of conversations that fulfill these criteria. UPHELD contains over 30,000 high-quality human-written turns of dialogue across hundreds of novel conversations, and for each turn we present multiple human labels along important dialogue coherence criteria like stylistic and content consistency. Unlike datasets extracted from noisy sources \cite{lison-tiedemann-2016-opensubtitles2016} or those relying on crowdsourced platforms where annotator backgrounds are unknown \cite{Zhang2018PersonaChat}, UPHELD utilized \textbf{professional script writers} to guarantee high conversational quality, which has been shown to lead to notable improvements in the quality and authenticity of dataset collection \cite{Pilan2024Conversational, Yin2024Quality}.

Our strategy for data collection and general approach allows UPHELD clear divergence from pre-existing dialogue benchmarks in three fundamental ways (a more detailed point-by-point overview is given in Appendix \ref{appendix:table}):

\begin{itemize}

    \item \textbf{Task-oriented, natural dialogue:} UPHELD focuses on task-oriented dialogues (e.g., math tutoring or trip planning), but all professional script writers were instructed to write dialogues in a natural, casual way. This deviates significantly from standard task-oriented datasets where agents follow a rigid ``set'' target or are constrained by encoded model knowledge. By providing writers the freedom to determine how the task is navigated, we capture a more naturalistic intersection of utility and human-like interaction.

     \item \textbf{Realistic conversational turn density:} While some datasets focus on extremely long conversations, such as interviews or LoCMo \cite{Wang2024LoCMo}, most human-to-human interactions are more concise and especially in the guided task-oriented space. Our analysis of human-to-human interaction in available datasets shows that real-world dialogue rarely exceeds 20 turns. Appendix~\ref{appendix:table} shows that the average of fully human-collected datasets is ~9.3 turns, while the Topical-Chat dataset \cite{topical} maxes at 21 turns. Our writers were given the freedom to determine conversation lengths and end at natural stopping points to keep the interactions as natural as possible, which means UPHELD better reflects real-world turn distributions.

    \item \textbf{Emphasis on conversational fluency versus information retrieval:} We deliberately avoid tasks requiring complex context, such as complex mathematics or coding. Instead, UPHELD focuses on the model’s ability to hold a coherent and reasonable conversation throughout an entire dialogue, and in service of everyday tasks that do not require heavy context engineering. This distinguishes our work from benchmarks like Wizard of Wikipedia \cite{Dinan2018WoW}, which prioritize knowledge-grounding over pure conversational dynamics.

\end{itemize}
 
In order to further illustrate the critical gaps that UPHELD was designed to bridge, we include a deep analysis of existing methods like LLM-as-a-judge or classical NLP metrics, and where they fall short on evaluating LLMs in conversational settings when tested against UPHELD. Our analysis shows that these metrics perform poorly for the task in question, and we demonstrate how to use UPHELD to generate better evaluation metrics through a simple machine-learned mixture-of-judges (ensemble) metric. Across evaluations, the proposed metrics demonstrate higher correlation with human judgments than single LLM-as-a-judge metrics as well as established syntactic and semantic scoring methods.

Our main contributions are as follows:
 \begin{itemize}
  \item We present UPHELD, a novel dataset of 10,000+ dialogue turns crafted by \textbf{professional script writers}. With \textbf{153,578 dense, per-turn human labels} that rate turn quality along various coherence criteria, UPHELD offers an evaluative scale nearly 10x that of comparable human-verified benchmarks \cite{Zhang2023Comprehensive}, effectively bridging the quality and volume relative to crowdsourced or synthetic datasets like in \cite{Yin2024Quality, Rahmani2025Biases}.
  \item We focus on realistic task-oriented dialogues tied to natural dialogue length distributions rather than purely casual, open-ended dialogue. In doing so, UPHELD effectively evaluates functional utility within naturalistic human cognitive boundaries, avoiding the artificial turn-length inflation common in synthetic datasets \cite{Wang2025Limits}.
  \item We introduce a multi-turn consistency metric that evaluates dialogue through turn-level coherence and reasonableness. We demonstrate that while standard metrics exhibit a significant performance gap on this task, our proposed Mixture-of-Judges ensemble metric is more successful at predicting expert human judgment by $\sim$30\% over individual LLM judges.
\end{itemize}

\section{Related Work}

Evaluating the performance of large language models (LLMs) is a critical area of research. Current benchmarks, such as those on the Open LLM leaderboard (now archived) \citep{llm-leader}, often focus on task-based evaluations using input-output metrics. These metrics include task-based generation evaluations like IFEval \citep{Ye2023IfEval} which evaluates the ability of LLMs to follow instructions and BBH (Big Bench Hard \citep{bbh}), which comprises 23 groups of tasks such as \textit{word sorting}, \textit{casual judgment}, \textit{navigate}. These and similar task-with-answer evaluations utilize metrics such as Exact Match (EM), Precision/Recall/F1, and substring-based accuracy. 
Specific examples include MATH for mathematical problems \citep{math}, and GPQA for multi-choice question answering \citep{GPQA}.  Unlike benchmarks focused on specific tasks (MATH, GPQA) or conversation quality, HELM \citep{liang2023helm} provides a unified framework evaluating models across different scenarios and metrics, emphasizing evaluation methodology rather than a specific task domain. However, it is still single turn based and is built on top of right/wrong factual answer comparison.

Question answering (QA) benchmarks are also prevalent. MUSR \citep{musr} is a narrative-based QA evaluation where the input is a paragraph and the output is an answer with an evaluation score. MMLU-PRO \citep{Wang2024MMLUPro} combines task-based evaluation with QA and chain-of-thought prompting. Recently, using an LLM-as-a-judge for evaluation is another approach which enables evaluating without having a ground truth or a reference answer \citep{zheng2023judgingllmasajudgemtbenchchatbot} \citep{Zhang2023Comprehensive}, \citep{botchat}, \citep{qa-correctness}.
While the aforementioned benchmarks are prominent, several related areas of work are not typically included in leaderboards focused on single-turn evaluations, particularly those concerning longer conversations. These include approaches similar to Chatbot Arena \citep{chiang2024chatbot}, MultiHop QA \citep{multihop} which involves answering questions over multiple documents or turns, and MT-eval which focuses on evaluating conversations directly and multi-dimensionally \citep{mt-eval}. BotChat \citep{botchat} is also noted as a method for conversation generation evaluation. 

The above works often rely on input-output pairs, exact matches, or scores tied to specific correct answers, which are well-suited for evaluating the ability to perform discrete tasks or extract factual information. 
However, they may not be appropriate for evaluating long, non-topical conversations. Such interactions involve sustained coherence, context management over many turns, engaging dialogue, and the ability to handle subjective or open-ended discussions that do not have a clear and concise "correct" answer.
Current human-scale conversation datasets focusing on non-factual conversations, including MuTal \citep{mu-tal}, Topical-chat \citep{topical}, LLM-arena \citep{zheng2023judgingllmasajudgemtbenchchatbot}, and DailyDialogue (Multi-turn) \citep{daily-dialogue}, often lack authentic human-to-human interaction or are constrained to predefined topics. Recognizing these limitations, and to construct a robust benchmark, we identified the need for a dataset characterized by comprehensive curation, expert authorship, and human annotation. This paper introduces such a dataset, which we have meticulously collected and organized, with details provided in the ensuing sections.

\section{Datasets}

\subsection{UPHELD dataset} \label{upwork_dataset}

We first collected conversational data by tasking a panel of professional writers hired from the Upwork freelancer marketplace to create open-ended, naturalistic dialogues across a diverse range of topics like customer service and education. The writers were explicitly instructed to create longer dialogues characterized by conversational complexity and exploratory interaction, rather than producing conversations with predetermined or narrow outcomes. For example, the writers developed dialogues exploring scenarios such as mobile phone selection, travel planning, or problem-solving discussions that were goal-oriented and where multiple nuanced exchanges would be appropriate. Writers were also instructed to avoid overly scripted or linear communication patterns. This approach ensures that the dataset captures the diverse and intricate nonlinear nature of human conversation. We also took steps to alleviate any potential writer bias (see details in Appendix \ref{appendix:bias}). The UPHELD dataset is publicly available under a permissive  \url{https://creativecommons.org/licenses/by/4.0/}{CC-BY-4.0 license} and downloadable from \url{https://github.com/upwork/upwork-ai-research/tree/master/data/upheld}.

\subsubsection{Input data}

Given our initial set of rich natural language conversations, various LLM models were then used to output candidate completions at every turn of every conversation. Specifically, models were presented with conversation history up to a specific point, with the next human-written turn withheld. Models then generated a predicted next turn. The following models were used to generate the dataset:
a) \textbf{GPT-3.5} \citep{gpt-3.5}; b) \textbf{GPT-4o} \citep{gpt-4o}; c) \textbf{GPT-5} \citep{gpt-5}; d) \textbf{Llama-3.1-70b} and \textbf{Llama-3.3-70b} \citep{llama3.1}; e) \textbf{Kimi-K2.5} \citep{kimik2}; f) \textbf{Qwen3-32B} \citep{qwen3}; g) \textbf{DeepSeek v3} \citep{deepsek} ) \textbf{custom model} we call \textbf{Uma-V0.5} (a fine-tuned version of Llama-3.1: see Section \ref{sec:uma}). In addition to these, we added two baseline candidate completions: a) a version of the ground truth (human) output rephrased using GPT-4o and b) a random completion. These additional baselines provide an important calibration signal for the dataset, and as expected we found that the rephrased dataset received the highest marks while the random dataset received the lowest marks when evaluated by a human labeler. These checks provide us with additional confidence that humans are fair judges of our evaluation axes -- style, content, and reasonableness. 

At every possible conversational turn, annotators were then given (1) the chat history up to that point, (2) the ground truth human completion (Option A), and (3) one of the LLM-generated candidate completions (Option B). The provenance of options A and B were not disclosed to the annotators. Their task was to compare these continuations on content consistency, style consistency, and general reasonableness. The tasks, metric scales, and instructions were fine-tuned through two paid pilot studies. Example data-points within UPHELD are given in Table \ref{tbl:examples}, and full instructions provided to the annotators can be found in Appendix \ref{appendix:graph}.

\begin{table*}[t]
\caption{Input conversations and associated annotator ratings from a single annotator (all scores are out of 5). More examples can be found in Appendix~\ref{appendix:examples}.}
\centering
\small
\setlength{\tabcolsep}{4pt}
\renewcommand{\arraystretch}{1.25}

\begin{tabularx}{\textwidth}{p{0.28\textwidth} X X}
\toprule
\textbf{chat\_history} &
\textit{user}: I'm shopping for a new car and can't decide what to get. \newline
\textit{assistant}: I can help with that. What size car are you looking for? &
\textit{user}: I'm really frustrated. I've been trying to get this printer to work all morning and it just won't print anything! \newline
\textit{assistant}: I understand that must be really annoying. Let’s work together to fix this. Can you tell me if you see any error messages on your printer or computer? \newline
\textit{user}: No error messages. It just sits there doing nothing whenever I try to print! \newline
\textit{assistant}: Alright, let's try a few steps to see if we can get it responding. First, could you check whether the printer is set as the default printer on your computer? \newline
\textit{user}: How do I check that?
\\
\midrule
\textbf{Option A} &
The Ford F-150 is one of the top selling trucks in the country. Have you taken one of those for a test drive? &
If you’re using a Windows PC, you can go to 'Devices and Printers' in your control panel. For Mac, you can check under 'System Preferences' and then 'Printers \& Scanners'.
\\
\midrule
\textbf{Option B} &
Trucks are great for hauling and towing. Here are a few questions to help narrow down your options: \newline
1. What will you be using the truck for most often? (e.g., work, off-roading, daily driving, towing a trailer) \newline
2. How many passengers will the truck need to seat? \newline
3. Do you have a preferred fuel type (gasoline, diesel, hybrid)? \newline
4. What's your budget for the truck? \newline
5. Do you have a specific brand or model in mind? \newline
6. Do you need any specific features, such as 4-wheel drive, a certain bed size, or advanced safety features
&
It's easy. On a Windows computer, you can go to the Control Panel, then Devices and Printers (or Printers and Devices in some versions). Look for your printer's name and see if it has a checkmark next to it that says ``Default printer''. If it doesn't, right-click on the printer and select ``Set as default printer''. \newline
On a Mac, you can go to System Preferences, then Printers \& Scanners. Look for your printer's name and make sure it's selected as the default printer. \newline
If you're still having trouble, we can try some other steps. Have you tried restarting the printer
\\
\midrule
\textbf{score\_task\_1\_content} &
1: Strongly Disagree -- no content overlap &
3: Neutral -- some overlap of content
\\
\textbf{score\_task\_2\_style} &
1: Different styles &
1: Different styles
\\
\textbf{score\_task\_3\_reasonableness} &
1: Not a reasonable continuation &
5: A reasonable continuation
\\
\bottomrule
\end{tabularx}

\label{tbl:examples}
\end{table*}

\subsubsection{Annotation dimensions}
Across all candidate models, each model-generated turn was evaluated by five independent human annotators using the following scoring criteria (see full instructions in Appendix \ref{appendix:graph}). 
\begin{itemize}
    \item Content Equivalence: A 5-point Likert scale (1-5) assessing semantic similarity between generated and reference turns.
    \item Style Equivalence: A 3-point Likert scale (1-3) measuring linguistic and stylistic correspondence.
    \item Utterance Reasonableness: A binary scale (1: unreasonable, 5: reasonable) evaluating the contextual appropriateness and coherence of the generated response,
\end{itemize}

The Style Equivalence scale we used is on a three level scale with scores (1 - 3 - 5) based on the initial pilot studies were participants suggested that 1-5 "standard" scale for "Style Equivalence" was more confusing compared to "Content Equivalence"  The annotators were provided with the explanation of these three levels as well. For the reasonableness scale we explored different options including non-binary and  pairwise comparisons (e.g., "Answer A is more reasonable than Answer B"). Based on the same pilot study we opted for a binary ``Reasonableness''. During this phase, we found that because reasonableness is often perceived by humans as an absolute quality—either a response makes sense within the dialogue flow or it does not—annotators struggled to consistently rank one reasonable response over another. Our conclusion was that if humans primarily evaluate reasonableness as a binary attribute, forcing them to compare two such responses introduces significant subjective noise and reduces inter-annotator agreement.

\subsubsection{Data statistics}

In total we collected complete evaluation labels for 800 conversations made up out of 9,203 turns/utterances, with with an average of 11.29 turns per conversation collected for this study and additional conversations from existing datasets \citep{mu-tal} to serve as control points. Each predicted turn was evaluated by five human annotators, and each annotator judge labeled at most 2,000 conversation continuations. Overall, we generated 30,715 sets of labels, or 153,578 labels (one set was annotated by 5 annotators). The ground truth conversations consist of 5.6 turn pairs (user-assistant) or 11.29 dialogue turns on average. The average length of the conversation history annotators analyzed was 560 characters and the length of the judged turns was on average 245 characters.

\subsubsection{Dataset content and quality checks}

The dataset was curated to include a wide range of topics and situations. The writers were given high-level scenario descriptions and stylistic guidelines and were instructed to produce naturalistic, human-like conversations without being constrained to specific task templates. While some conversations contain elements of reasoning or factual exchange, these emerged organically rather than being explicitly assigned tasks. The dialogues in UPHELD cover a diverse range of human-scale situational contexts, including:

\begin{itemize}
    \item Decision Making: (e.g., A gym owner helping a client decide on a personal trainer).
    \item Customer Service: (e.g., Navigating a website to cancel a subscription service).
    \item General Life Advice: (e.g., An academic advisor helping a student choose a major).
    \item Tutoring: (e.g., Solving multi-step math word problems involving currency and change).
    \item Task Assistance: (e.g., A chef guiding a novice through a recipe for the first time).
    \item Debate: (e.g., Speculative discussion on the state of the world in 50 years).

\end{itemize}

To minimize bias and ensure high quality of the collected data, we added several layers of checks and run several paid pilot studies (see Appendix~\ref{appendix:bias}).
\begin{itemize}
    \item \emph{Writer Selection}: All writers possessed high success rates on Upwork and passed a rigorous paid pilot phase evaluated by professional UX researchers for diversity and guideline adherence. 
    \item \emph{Diverse Expertise}: Writers were recruited from varied backgrounds (novel writing, education, copywriting) to ensure situational and stylistic diversity.
    \item {Multi-Stage Review}: Each conversation underwent independent checks by professional proofreaders for coherence, and naturalness and expert peer review for stylistic diversity.
    \item \emph{Enforcement:} Low-quality, inconsistent, or repetitive conversations were actively filtered and revised; each dialogue was reviewed by multiple independent professionals (typically 2-3 reviewers per conversation).
\end{itemize}

All writers were recruited via Upwork under transparent contracts and compensated at professional market rates. They were explicitly informed that their work would be used for AI research and provided informed consent. No personally identifiable information (PII) is included in the dataset.  All annotators were recruited via an Upwork as a part of an agency and were compensated at a professional market rate including a paid pilot study to calibrate the amount of work and fair compensation.

\subsubsection{Verification datasets} \label{verification_datasets}

To further validate our findings, we construct additional verification datasets by augmenting LLM-Arena \cite{llm-leader} and Topical-Chat \citep{topical}. The overall procedure consisted of three steps: (i) deriving a single ``ground truth answer'' from each data point of each existing dataset (see below), (ii) generating an alternative continuation with GPT-4o, and (iii) collecting human judgments following the UPHELD annotation protocol. We include all additional verification labels within our dataset for reproducibility.

Derivation of Ground Truth:
    \begin{itemize}
   
    \item[a.] LLM-Arena:
    \begin{itemize}
        \item  Data Point Description: each dialogue is accompanied by two model-generated continuations plus a human preference label.
        \item Ground Truth Extraction: for every conversation we enumerated all candidate pairs, tallied human preferences, and chose the majority-preferred continuation as the reference (provided a clear winner existed).
    \end{itemize}
    \item[b.] Topical-Chat:
    \begin{itemize}
        \item Data Point Description: human‐to‐human dialogues grounded in specific topic selections.
        \item Ground Truth Extraction: to mitigate cold-start artifacts, we extracted segments spanning turns 5–7. We then treated turns 1–5 (or 1–7) as the model input and selected the next human turn (turn 6 or 8) as the ground truth continuation.
        
    \end{itemize}
    \end{itemize} 

After standard quality control (i.e. filtering for missing data and badly formatted inputs), we obtained 12,305 pairwise preference judgments. We note that both these verification datasets, although useful for verification, are still relatively deficient in freeform human-to-human interaction and focus on a limited set of pre-defined topics. As such they should be treated as verification datasets only and not as valid replacements for UPHELD.

\section{Metrics}
\label{sec:metrics}

In total, 12 candidate metrics were assessed for their ability to evaluate longer conversations via correlation with UPHELD labels. These metrics were grouped into 3 distinct groups: 1) token-based -- metrics quantifying similarity based on exact overlap of tokens 2) semantic-based -- metrics quantifying similarity based on semantic overlap (e.g. embedding models); and 3) LLM-based -- metrics employing some form of the LLM-as-a-judge paradigm.

Recall-Oriented Understudy for Gisting Evaluation (ROUGE \citep{rouge}) is a set of standard language metrics that compare automatically produced summaries or translations against a set of reference summaries or translations. Specifically, ROUGE-N measures the overlap of n-grams between the system-generated text and the reference text. ROUGE-L measures the longest common subsequence, which accounts for sentence-level structure similarity. 

We also explored cosine similarity between message embeddings as a measure of semantic similarity between the generated text and reference text. This approach is rooted in the work by \citet{bert} on Sentence-BERT embeddings, which have shown effectiveness in capturing semantic similarities in text data. BERTScore leverages the pre-trained contextual embeddings from BERT to evaluate text generation by matching words in candidate and reference sentences. It computes precision, recall, and F1 score, providing a more nuanced evaluation than traditional n-gram-based metrics.\citet{zhang_bert} introduced BERTScore as a robust metric for evaluating generated text.

We also tested LLM-as-a-judge metrics \citep{zheng2023judgingllmasajudgemtbenchchatbot} against UPHELD. This approach involves using a separate LLM to score the outputs based on various criteria, such as coherence, relevance, and overall quality. We used both binary (yes/no) and Likert scale (1-5) judgments, with and without explanations. Prompts for judges can be found in Appendix \ref{appendix:prompts}. In addition to these LLM-judge metrics, we also ran experiments using the same prompts as those given to human raters (Appendix \ref{appendix:prompts}) and observed lower scores than with independently created LLM-judge prompts. 

\subsection{Reference-free vs human-grounded metrics}

Fundamentally, UPHELD uses a ground truth reference to generate our human labels. In contrast, reference-free evaluation of LLMs \citet{g-eval} relies on human preference, and LLM outputs can reliably reproduce these preferences, indicating their performance is consistent with human judgments \citep{Zhang2023Comprehensive}. However, reference-free preference datasets also incur significant limitations, such as poor performance at judging long task-oriented dialogues and weakened judgment reliability on out-of-distribution tasks \citep{no-free-ref}. UPHELD is designed specifically to tackle these limitations by adopting a multi-prong approach via reference-full content similarity annotations,while still enabling reference-free evaluation through the reasonableness annotations and implicit stylistic evaluation through the style annotations.

The reference-full approach raises a question around dialogue multiplicity: a single input might incur multiple valid outputs, so how are we sure our ground truths are well defined? UPHELD's design reduces susceptibility to this issue in two ways: (1) two of our key label categories (style and reasonableness) are well-defined even with dialogue multiplicity, and (2) UPHELD dialogues primarily revolve around task-oriented settings, which means content accuracy is a well-defined metric. For example, while opinion-oriented conversations (e.g. \textit{Who makes the best Caesar salad?}) are susceptible to dialogue multiplicity, our task-oriented dialogues (e.g. \textit{How to make a Caesar salad?}) are not. To quantify this effect, we ran the following experiment exploring multiplicity.

\subsection {UPHELD and Conversational Multiplicity} \label{appendix:multi-experminet}

A valid concern may be that direct comparisons to a reference human answer may be inappropriate in settings when a particular prefix can lead to a multiplicity of valid responses. This effect may be prevalent especially when the prefix is asking for an opinion (e.g. ``What is your favorite animal?"). We, however, observe that UPHELD dialogues avoid this potential pitfall as they are not strictly freeform, but are all targeted towards completion of a specific well-defined task. In this context, there is some notion of correct ground truth, and we specifically hired professionals who are experts at these tasks (see Appendix \ref{appendix:bias}). To put it simply, our task setting is analogous to the difference between \textit{what kind of salads do you like?} (which has ambiguity and dialogue multiplicity) and \textit{how do I make a Caesar salad?} (which is much more constrained and has a more well-defined ground truth). To quantify this, we generated 100 open-ended questions (GPT-4o) and then generated two possible completions with GPT-4o at moderately high temperature ($\tau=1$) to those questions. We did the same with 100 UPHELD turn completions. We then asked GPT-4o to judge whether the two possible completions contain similar content. The results are as follows:

\begin{table}[htbp]
\small
    \centering
        \caption{Semantic Consistency Performance across Different Datasets}

    \begin{tabular}{lc}
        \hline
        Dataset & Semantic Consistency (\%) \\
        \hline
        Open-Ended & 74 \\
        UPHELD (first turn only) & 93 \\
        UPHELD (random turn) & 93 \\
        \hline
    \end{tabular}
    \label{tab:semantic_consistency}
\end{table}

UPHELD exhibits much higher semantic consistency in the output, which means that UPHELD dialogues admit much less conversational multiplicity than more freeform datasets. The results show that UPHELD dialogues admit significantly higher output consistency (93\%) compared to freeform dialogues (74\%), demonstrating that our reference-full approach still allows us to collect meaningful labels on ground truth content overlap. This supports our hypothesis that targeted task-focused conversations like those in UPHELD admit well-defined "ground truth" references. Interestingly, UPHELD maintains high output consistency even when we only analyze the first turn, which is where we would expect more branching/multiplicity during a dialogue. Note that these results are likely an underestimate of the true consistency, since sampling multiple LLM outputs would induce additional randomness that likely would not exist within natural human dialogue.

\section{Experiments}\label{sec:experiments}

To demonstrate the value of human-scale long conversational evaluation, we present a series of experiments showing that (a) UPHELD base conversations substantially improve LLM conversational fidelity, and (b) naive evaluation metrics degrade on human-scale long dialogues, motivating development of simple ensemble metrics that outperform baseline approaches. These metrics also perform well on our validation datasets (Topical and LLM Arena), showing that a method developed with UPHELD is transferrable to other contexts. We also include discussion on user disagreement within UPHELD. 

\subsection{UPHELD as a Fine-Tuning Dataset}\label{sec:uma}

\begin{figure*}
    \centering
    \includegraphics[width=0.85\linewidth]{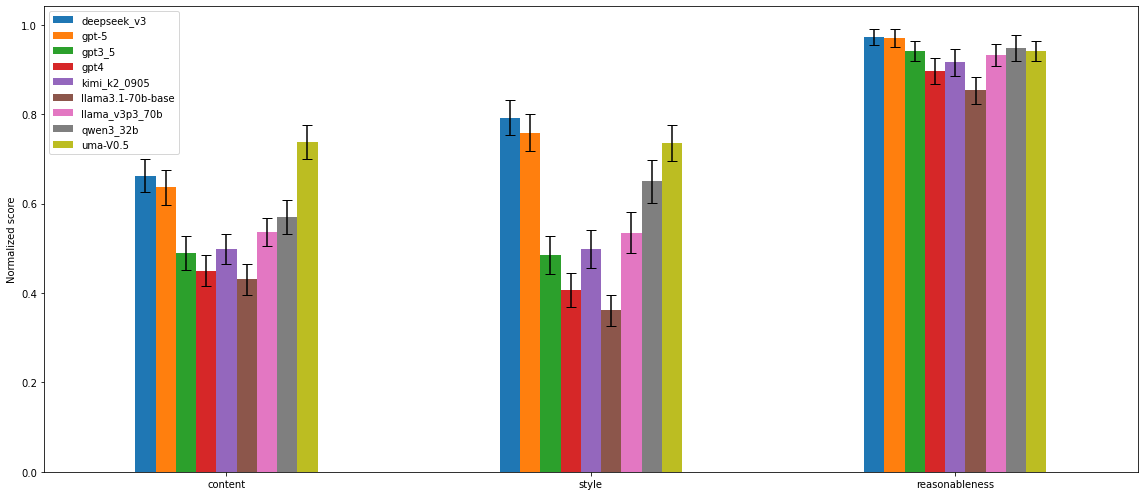}
    \caption{Aggregate human labeler scores as a share of the perfect score (see Appendix \ref{appendix:aggregate}) for each model on the UPHELD dataset within each label category. The custom model, which was fine-tuned on top of Llama3.1-70b using a held-out set of training data within the UPHELD dataset, performs significantly better than its baseline model rivaling or beating the SOTA model and demonstrating that UPHELD contains information that can greatly improve conversational quality of base models.}
    \label{fig:scores}
\end{figure*}

For our experiments we report the numbers on a subset of UPHELD consisting of 756 label sets spread across conversations, each labeled by up to five annotators. The total number of unique utterances were 4{,}777 averaging at 6.3 pairs of user-assistant turns. The remaining conversations were used exclusively for fine-tuning and are therefore excluded from evaluation to avoid data leakage. An important way to validate the quality of data within the UPHELD dataset is to assess how the UPHELD scores differ between the \textit{custom} model, which was fine-tuned on the base data, and the other baseline models. To do so, we directly plot the mean content, style, and reasonableness scores within the UPHELD dataset for the different models in Figure \ref{fig:scores}. Older base models (GPT-3.5, GPT-4o, and Llama3.1-70b-base) exhibit lower performance in both content accuracy and style accuracy. This indicates a tendency for these models to deviate from the intended conversational style and introduce content inconsistencies. Specifically, the base models demonstrate a substantial deficit in maintaining the stylistic integrity and topical coherence of the conversation, diverging from the trajectory established by human writers. 

In contrast, the custom model, fine-tuned for extended dialogue on a held-out set, shows a marked improvement and achieves approximately a 40\% increase in both content and style accuracy compared to the base models and reaches comparable scores to newer, best-in class, SOTA models like GPT-5 and Deepseek v3. The trained model achieves this while retaining the core functionality of its model base (Llama-3.1-70b), which we validated at test time and also at training time where we observed minimal overfitting within loss curves.

Reasonableness is fairly flat across all models, which is not surprising as LLMs tend to output reasonable results regardless of style or content consistency. These results highlight that the UPHELD dataset's conversations are both learnable and encode behavior that is not well-exposed within these models' pre-training datasets. our decision to use a single-model judge for "Reasonableness" was intended to specifically test the performance of reference-free versus reference-full evaluation. The resulting low variance between model scores serves as a strong signal that reference-free metrics currently struggle to capture the nuances of human-scale dialogue. This outcome reinforces our thesis that a reference-full approach, supported by high-quality human data like UPHELD, remains a vital and more reliable methodology for the field. 

\subsection{Evaluation Metric Performance on UPHELD}

Given that UPHELD is designed to help develop novel evaluation metrics for long conversations, it is instructive to see how this development works in practice. We start this section with an analysis on how traditional metrics perform poorly on UPHELD, and how simple modifications to the traditional metrics provide a significant boost in performance. All results in this section are presented as 5-fold cross validation results on a 20\% held-out set. For numerical values, we calculated Pearson correlation between the scores and human judges; for categorical metrics we calculated Cramér's V correlation coefficients; and for binary metrics we report point-biserial correlation coefficient. 

\subsubsection{Traditional Metrics}
\begin{table}[h]
\caption{Results on the UPHELD datasets for various candidate evaluation metrics on the first evaluation task (content accuracy). Ensemble metrics clearly perform better for both the UPHELD dataset and both verification datasets.}
\centering
\small
\setlength{\tabcolsep}{5pt}
\renewcommand{\arraystretch}{1.15}

\resizebox{\columnwidth}{!}{%
\begin{tabular}{lccc}
\toprule
\textbf{Metric} & \textbf{UPHELD} & \textbf{LLM Arena} & \textbf{Topical-Chat} \\
\midrule
\multicolumn{4}{l}{\textbf{Semantic Metrics}} \\
message\_embedding\_cos\_sim
& 0.58$\pm$\tiny{0.003}
& 0.41$\pm$\tiny{0.004}
& 0.40$\pm$\tiny{0.008} \\
bert\_score\_precision
& 0.36$\pm$\tiny{0.005}
& 0.39$\pm$\tiny{0.004}
& 0.32$\pm$\tiny{0.008} \\
bert\_score\_recall
& 0.40$\pm$\tiny{0.005}
& 0.42$\pm$\tiny{0.004}
& 0.40$\pm$\tiny{0.008} \\
bert\_score\_F1
& 0.40$\pm$\tiny{0.005}
& 0.45$\pm$\tiny{0.003}
& 0.42$\pm$\tiny{0.008} \\
\midrule
\multicolumn{4}{l}{\textbf{Token-based Metrics}} \\
rouge1
& 0.34$\pm$\tiny{0.005}
& 0.49$\pm$\tiny{0.003}
& 0.33$\pm$\tiny{0.009} \\
rouge2
& 0.24$\pm$\tiny{0.005}
& 0.43$\pm$\tiny{0.003}
& 0.24$\pm$\tiny{0.009} \\
rougeL
& 0.32$\pm$\tiny{0.005}
& 0.41$\pm$\tiny{0.004}
& 0.34$\pm$\tiny{0.010} \\
rougeLsum
& 0.31$\pm$\tiny{0.005}
& 0.50$\pm$\tiny{0.003}
& 0.34$\pm$\tiny{0.009} \\
\midrule
\multicolumn{4}{l}{\textbf{LLM-as-a-judge Metrics}} \\
llm\_judge\_yes\_no
& 0.40$\pm$\tiny{0.012}
& 0.32$\pm$\tiny{0.015}
& 0.57$\pm$\tiny{0.016} \\
llm\_judge\_yes\_no\_explain
& 0.23$\pm$\tiny{0.018}
& 0.32$\pm$\tiny{0.014}
& 0.29$\pm$\tiny{0.013} \\
llm\_judge\_likert\_1\_5
& 0.26$\pm$\tiny{0.008}
& 0.41$\pm$\tiny{0.011}
& 0.39$\pm$\tiny{0.014} \\
llm\_judge\_likert\_1\_5\_explain
& 0.28$\pm$\tiny{0.009}
& 0.41$\pm$\tiny{0.013}
& 0.34$\pm$\tiny{0.013} \\
\midrule
\multicolumn{4}{l}{\textbf{Ensembled ML Metrics (Ours)}} \\
Linear Regression
& 0.59$\pm$\tiny{0.004}
& \textbf{0.50}$\pm$\tiny{0.003}
& \textbf{0.58}$\pm$\tiny{0.006} \\
SVM
& 0.55$\pm$\tiny{0.005}
& 0.49$\pm$\tiny{0.003}
& 0.54$\pm$\tiny{0.008} \\
Random Forest
& \textbf{0.63}$\pm$\tiny{0.004}
& 0.23$\pm$\tiny{0.004}
& 0.32$\pm$\tiny{0.009} \\
\bottomrule
\end{tabular}%
}
\label{tab:metric_comparison_content}
\end{table}

\begin{table}[h]
\caption{Results on the UPHELD datasets for various candidate evaluation metrics on the second evaluation task (style accuracy).}
\centering
\small
\setlength{\tabcolsep}{5pt}
\renewcommand{\arraystretch}{1.15}

\resizebox{\columnwidth}{!}{%
\begin{tabular}{lccc}
\toprule
\textbf{Metric} & \textbf{UPHELD} & \textbf{LLM Arena} & \textbf{Topical-Chat} \\
\midrule
\multicolumn{4}{l}{\textbf{Semantic Metrics}} \\
message\_embedding\_cos\_sim
& 0.41$\pm$\tiny{0.001}
& 0.42\tiny{$\pm$0.004}
& 0.29\tiny{$\pm$0.009} \\
bert\_score\_precision
& 0.37$\pm$\tiny{0.005}
& 0.39\tiny{$\pm$0.004}
& 0.36\tiny{$\pm$0.008} \\
bert\_score\_recall
& 0.36$\pm$\tiny{0.005}
& 0.50\tiny{$\pm$0.004}
& 0.27\tiny{$\pm$0.009} \\
bert\_score\_F1
& 0.41$\pm$\tiny{0.005}
& 0.49\tiny{$\pm$0.003}
& 0.35\tiny{$\pm$0.008} \\
\midrule
\multicolumn{4}{l}{\textbf{Token-based Metrics}} \\
rouge1
& 0.32$\pm$\tiny{0.005}
& 0.51$\pm$\tiny{0.003}
& 0.31$\pm$\tiny{0.010} \\
rouge2
& 0.20$\pm$\tiny{0.005}
& 0.43$\pm$\tiny{0.004}
& 0.16$\pm$\tiny{0.010} \\
rougeL
& 0.29$\pm$\tiny{0.005}
& 0.48$\pm$\tiny{0.004}
& 0.30$\pm$\tiny{0.009} \\
rougeLsum
& 0.29$\pm$\tiny{0.005}
& 0.48$\pm$\tiny{0.003}
& 0.30$\pm$\tiny{0.009} \\
\midrule
\multicolumn{4}{l}{\textbf{LLM-as-a-judge Metrics}} \\
llm\_judge\_yes\_no
& 0.35$\pm$\tiny{0.012}
& 0.08$\pm$\tiny{0.010}
& 0.36$\pm$\tiny{0.016} \\
llm\_judge\_yes\_no\_explain
& 0.15$\pm$\tiny{0.015}
& 0.08$\pm$\tiny{0.010}
& 0.24$\pm$\tiny{0.013} \\
llm\_judge\_likert\_1\_5
& 0.24$\pm$\tiny{0.010}
& 0.07$\pm$\tiny{0.009}
& 0.22$\pm$\tiny{0.014} \\
llm\_judge\_likert\_1\_5\_explain
& 0.20$\pm$\tiny{0.012}
& 0.05$\pm$\tiny{0.008}
& 0.20$\pm$\tiny{0.013} \\
\midrule
\multicolumn{4}{l}{\textbf{Ensembled ML Metrics (Ours)}} \\
Linear Regression
& 0.50$\pm$\tiny{0.004}
& 0.28$\pm$\tiny{0.004}
& 0.32$\pm$\tiny{0.008} \\
SVM
& 0.49$\pm$\tiny{0.005}
& \textbf{0.54}$\pm$\tiny{0.003}
& \textbf{0.55}$\pm$\tiny{0.009} \\
Random Forest
& \textbf{0.61}$\pm$\tiny{0.004}
& 0.04$\pm$\tiny{0.004}
& 0.22$\pm$\tiny{0.009} \\
\bottomrule
\end{tabular}%
}

\label{tab:metric_comparison_style}
\end{table}

The main results for traditional metrics (as defined in Section \ref{sec:metrics}) are shown in the first three rows of Tables \ref{tab:metric_comparison_content} and \ref{tab:metric_comparison_style}. The results reveal a weak to moderate correlation between metrics and human ratings. This suggests that no single metric captures the nuances of human assessment well within the UPHELD dataset. All definitions for metrics can be found in the Appendix \ref{appendix:metrics}. Interestingly, the semantic metrics (like bert\textunderscore score\textunderscore F1) demonstrated the highest correlation with human judgments for both content and style aspects. The token-based metrics showed stronger correlations than LLM-as-a-judge as well. These findings suggest that LLM-as-a-judge, despite being increasingly explored in the literature (e.g., \citep{ma} \citep{zhang_bert}), is a weak evaluator of human-scale longer conversations. Despite the observed strength of the semantic metrics, the results for traditional metrics show weak to moderate performance on UPHELD. This implies that relying solely on any single traditional metric inadequately captures the complexities of content and style quality in conversations, and we demonstrate that ensemble metrics can bridge this gap.

\subsubsection{Ensemble Metrics}

We hypothesize that individual metrics attend to distinct facets of text quality, and so a learned ensemble will perform better on the UPHELD dataset. We train linear regression, SVM, and random forest models to predict human scores using the individual automatic metric scores as input features. Using this approach, we derive new hybrid metrics and assess their correlation with human judgments.

The last 3 rows of Tables \ref{tab:metric_comparison_content} and \ref{tab:metric_comparison_style} show that the learned ensemble metrics exhibit higher correlations with human scores compared to any single metric in isolation. Notably, the random forest model yields substantial improvements of 30-40\% relative to the best-performing individual metrics on UPHELD, but an SVM ensemble produces consistently higher correlations to human judges across all  datasets.  Note that the ensemble metrics were trained only on the UPHELD data and then applied to the LLM Arena and Topical-Chat data. Although the transferability of the random forest model was poor, the SVM and linear regression ensembles indicate that remarkably, trained metrics developed just on the UPHELD dataset have exceptional transferability to other out-of-domain datasets.

The success of the ensemble metrics likely stems from their ability to integrate diverse signals captured by the individual metrics, mirroring the multifaceted nature of human evaluation. These findings strongly suggest that within complex settings like human-scale dialogue, learning to ensemble multiple automatic metrics offers a promising avenue for developing evaluation frameworks that more closely align with human judgments than relying on a single metric.

\subsection{Discussion Around Annotator Agreement}
\label{sec:disagreements}

Because each conversation turn was independently labeled by five human annotators, we also analyze model responses that elicited stronger human-human agreements. We observed moderate levels of consensus at average Cohen's $\kappa=0.33$. As expected, human-human agreement is highest on the \textit{random} and \textit{gpt4$\textunderscore$rephrase} baselines confirming the quality of the annotations, while being consistent across all other models. Further analysis on labeler agreement can be found in Appendix \ref{appendix:agree}.

\begin{table}[t]
\caption{LLM-as-a-judge performance on the UPHELD dataset for different dataset splits, grouped by the level of human annotator agreement within each bucket.}
\centering
\small
\setlength{\tabcolsep}{4pt}
\renewcommand{\arraystretch}{1.15}

\begin{tabular}{lccccc}
\toprule
& \multicolumn{5}{c}{\textbf{LLM-as-a-judge}} \\
\cmidrule(lr){2-6}
\textbf{Agreement}
& \textbf{Human}
& \textbf{Likert}
& \textbf{(explain)}
& \textbf{Binary}
& \textbf{(explain)} \\
\midrule
Plurality
& 0.47 & 0.34 & 0.36 & 0.755 & 0.72 \\
Majority
& 0.69 & 0.38 & 0.39 & 0.80 & 0.78 \\
Full 
& 1.00 & 0.63 & 0.56 & 0.93 & 0.92 \\
\bottomrule
\end{tabular}

\label{tab:matches}
\end{table}

We observed that approximately 25\% of data points had full agreement across all 5 human judges. Otherwise, we bin the level of agreement as follows: agreement across 2 out of 5 labels represents a ``plurality," while agreement across 3 or 4 labels represents a ``majority." We further quantify how well an LLM-as-a-judge evaluator agrees with this winning score relative to the agreement bin. Table \ref{tab:matches} shows that LLM-as-a-judge performance is heavily correlated to agreement level amongst the human labelers. This result demonstrates that human-human disagreement is a valid measure of data difficulty, and this uncertainty signal present in UPHELD may be integral in further evaluation metric development.

\section{Conclusion}


In this work, we introduced UPHELD: a dataset designed to evaluate LLMs in human-scale conversational settings. We collected tens of thousands high-quality human-annotated labels on crucial consistency metrics within human-scale conversations, along with the high-quality conversations themselves. Analysis of existing evaluation metrics on UPHELD reveals that they do not effectively capture the nuances of human judgment for assessing conversational quality. We further demonstrated that simple-to-learn ensemble metrics result in substantially improved correlations with human evaluations. Taken in aggregate, our findings highlight the potential for developing robust evaluation frameworks that better align with human perceptions of effective conversation using UPHELD.

\clearpage

\section*{Impact statement}
This study provides a valuable new dataset UPHELD and associated insights into its utility. As is, we believe it is already of significant interest to the machine learning and language modeling community, but we also believe it will help researchers to use the dataset to investigate more conversational verticals, especially those that are particularly relevant to enterprise applications such as targeted customer service and dialogues around more technical topics. 

\section*{Acknowledgment}
We would like to thank the Upwork team members whose contributions made this research possible: Jonathan Shen for his measurable contributions to Upwork's internal LLM stack which enables easy and scalable LLM training and evaluation, as well as Brett Levert who managed recruitment and quality controls for writers and annotators used in collecting the Upheld dataset. 

\newpage

\bibliographystyle{icml2026}
\bibliography{references}

\clearpage
\appendix
\appendix
\onecolumn
\section{Related Work comparison table} \label{appendix:table}
This section formally defines the key dimensions and associated metrics used to characterize and compare dialogue benchmark datasets. These dimensions establish the \textbf{data provenance, scale, task focus, and evaluation rigor} of a benchmark, which are critical for assessing its suitability for training and evaluating large language models (LLMs).

\hrule
\subsection{Data Statistics}
These dimensions shown in Table \ref{tab:comp_part1_final} quantify the scale and density of the linguistic resources within the benchmark.

\subsubsection{Scale Metrics}
\begin{itemize}
    \item \textbf{Total Conversations:} Represents the absolute count of distinct conversational threads or sessions included in the dataset.
    \item \textbf{Total Utterances:} Indicates the total number of individual speech acts or turns produced by all participants across the entire dataset.
\end{itemize}

\subsubsection{Density Metric}
\begin{itemize}
    \item \textbf{Total Comparison Labels:} Denotes the aggregate number of specific human or machine judgments collected for the purpose of comparing the quality of model-generated responses (e.g., A/B test results, preference rankings, or score annotations). This metric is a key indicator of the dataset's utility for \textbf{pairwise evaluation} and reinforcement learning from human feedback (RLHF).
\end{itemize}

\hrule
\subsection{Collection Information}
This dimension pertains to the \textbf{data generation process} and the level of human involvement in its creation and quality control.

\subsubsection{Data Provenance}
\begin{itemize}
    \item \textbf{Human (is the data generated by a human):} A binary metric indicating whether the dialogue turns were initially composed by human actors (\textbf{yes}) or whether they were generated synthetically, typically by an LLM (\textbf{no}).
    \item \textbf{Direct (are the conversations directly obtained from the source):} A binary metric indicating whether the data was captured raw from its source (e.g., direct transcription, human writing) (\textbf{yes}) or whether it underwent substantial intermediate processing, such as speech-to-text conversion or LLM-based reformulation (\textbf{no}).
\end{itemize}

\subsubsection{Quality Control}
\begin{itemize}
    \item \textbf{Verified (is there human verification on the quality of the conversation):} A binary metric indicating whether a human performed explicit quality checks, validation, or post-hoc filtering on the conversational data to ensure coherence, safety, or adherence to task instructions (\textbf{yes}).
\end{itemize}

\hrule
\subsection{Task Information}
These dimensions shown in Table \ref{tab:comp_part2_final} describe the \textbf{functional objective} and \textbf{knowledge requirements} imposed by the dataset's dialogue prompts.

\subsubsection{Task Type}
\begin{itemize}
    \item \textbf{Factual QA:} The task is constrained to generating a single, verifiably correct answer to a question.
    \item \textbf{Open Dialogue (Chit-Chat):} The dialogue is non-goal-oriented, focused on maintaining conversational flow, persona consistency, and engaging interaction.
    \item \textbf{Summarization:} The task requires the model to produce a concise abstract or summary of provided source material.
    \item \textbf{Specialized Tasks:} The benchmark encompasses a heterogeneous set of specific, domain-specific tasks, such as mathematical problem solving, code generation, or complex instruction following.
    \item \textbf{Context Retrieval:} The primary conversational goal is to retrieve or locate a specific piece of information from a provided or externally accessible knowledge base.
\end{itemize}

\subsubsection{Context Utilization}
This sub-dimension describes how external information is introduced to support the dialogue.
\begin{itemize}
    \item \textbf{RAG (Retrieval-Augmented Generation):} Context is inserted into the conversation using an algorithmic retrieval system, typically based on vector similarity.
    \item \textbf{Grounded Knowledge:} Context is inserted into the conversation in a deterministic or pre-defined manner, ensuring the LLM has access to a specific piece of knowledge for its response.
    \item \textbf{None:} No external context is provided; the model relies solely on the preceding dialogue history and its parametric knowledge.
\end{itemize}

\subsubsection{Prediction Format}
\begin{itemize}
    \item \textbf{Is Conversation:} A binary metric indicating whether the dataset is entirely formatted as a sequence of alternating user/assistant turns (\textbf{yes}) or if it includes substantial non-dialogue components, such as initial instructions or descriptive context (\textbf{no}).
\end{itemize}

\hrule
\subsection{Results Verification}
The dimensions shown in Table \ref{tab:comp_part3_final} assess the \textbf{methodology and validity} of the benchmark's evaluation process.

\subsubsection{Evaluation Basis}
\begin{itemize}
    \item \textbf{Explicit:} Evaluation relies on a direct, measurable comparison of the model's output against a predefined, canonical "ground-truth" reference answer (e.g., F1, exact match).
    \item \textbf{Reference-Free:} Evaluation is performed without a canonical target answer, relying instead on subjective judgment, typically from LLM-as-a-Judge systems or human preference rankings.
\end{itemize}

\subsubsection{Annotation Quality and Transparency}
\begin{itemize}
    \item \textbf{Uses Human Annotators:} A binary metric indicating whether human annotators were employed to subjectively rate or score the quality of the model's generated outputs (\textbf{yes/no}).
    \item \textbf{Correlation with Human:} A binary metric indicating whether the benchmark report provides an explicit measure of the agreement (e.g., Pearson's $r$) between the reported automatic evaluation metrics and the parallel human evaluation results (\textbf{yes/no}).
    \item \textbf{Provides Author/Creator Selection Details:} A binary metric indicating whether the documentation includes detailed information regarding the qualifications, recruitment, or background of the human writers and annotators employed for data generation and quality evaluation (\textbf{yes/no}). This speaks directly to \textbf{data quality and reproducibility}.
\end{itemize}

\begin{table}[!htbp]
\caption{Comparison of Dataset Properties (Part 1: Data Statistics)}
\label{tab:comp_part1_final}
\centering
\small 
\setlength{\tabcolsep}{4pt}
\begin{tabularx}{\textwidth}{@{} >{\raggedright\arraybackslash}p{3.4cm} >{\raggedright\arraybackslash}p{4cm} c c >{\raggedright\arraybackslash}p{2.0cm} @{}}
\toprule
\textbf{Dataset/Benchmark} & \textbf{Link} & \textbf{Total Conv.} & \textbf{Avg Utt.} & \textbf{Comparisons} \\
\midrule
UPHELD & & 53 & 10.2 & 36,873 \\
\midrule
MuTal & \citep{mu-tal} & 6,731 & 4.73 & 6,371 \\
Topical-Chat & \citep{topical} & 9,058 & 21.9 & 150 (human) \\
LLM Arena (crowd-sourced) & \citep{zheng2023judgingllmasajudgemtbenchchatbot} & 33,000 & 1.2 & 33,000 \\
LLM Arena (annotated) & \citep{zheng2023judgingllmasajudgemtbenchchatbot} & 3,000 & 2 & 3,000 \\
DailyDialogue & \citep{daily-dialogue} & 13,118 & 7.9 & 11,118 \\
IFEval & \citep{Ye2023IfEval} & 250 & 2 & 250 \\
MT-bench & \citep{mt-eval} & 80 & 2 & 80 \\
MMLU-pro & \cite{Wang2024MMLUPro} & 12,032 & 2 & 12,032 \\
HopQA & \cite{multihop} & 7,405 & 2 & 7,405 \\
MT-bench-101 & \citep{Zheng2024MTBench101} & 1,388 & 3.03 & 1,388(100 human) \\
MultiChallenge & \citep{Sirdeshmukh2025MultiChallenge} & 273 & 5 & 273 \\
LoCMo & \citep{Wang2024LoCMo} & 50 & 304.9 & 50(?) \\
LongTimeNoSee & \citep{Xu2022LongTimenose} & 27,501 & 16.34 & 200 \\
COMPREHENSIVE & \citep{Zhang2023Comprehensive} & 2,030 & 13.3 & 2,030 \\
COMPREHENSIVE-turn & \cite{Zhang2023Comprehensive} & 417 & 7.4 & 3,901 \\
ChatEval & \citep{Chang2023ChatEval} & 80(open QA)+60(dialogue) & 2 & 440 \\
Persona-chat & \citep{Zhang2018PersonaChat} & 10,907(1000 test/100 human) & 14.85 & 15,602 (test) \\
Wizards of Wikipedia & \citep{Dinan2018WoW} & 18,430 & 9.05 & 166,787(300 human) \\
SODA & \citep{Kim2023SODA} & $\sim$1.3m & 7.6 & 300(human evaluated sample) \\
Empathetic Dialogues & \citep{rashkin-etal-2019-towards} & 25,000 & 4.0 & 25,000 \\
Blended Skill Talk & \citep{smith2020blendedskill} & 6,810 & 4.5 & 6,810 \\
\midrule
\multicolumn{5}{@{}l}{\textbf{plain datasets (no benchmarks)}} \\
\midrule
Ubuntu dialogue context & \citep{Lowe2015Ubuntu} & $\sim$1 million & 7.7 & n/a \\
Open Subtitles & \citep{lison-tiedemann-2016-opensubtitles2016} & 3.7 million & 16.75 (sentences) & n/a \\
Twitter-dataset & \citep{Ritter2011Twitter} & 4,323 & 2 & 2161 \\
\bottomrule
\end{tabularx}
\end{table}

\begin{table}[!htbp]
\caption{Comparison of Dataset Properties (Part 2: Collection \& Task Information)}
\label{tab:comp_part2_final}
\centering
\small
\setlength{\tabcolsep}{3pt}
\begin{tabularx}{\textwidth}{@{} >{\raggedright\arraybackslash}p{2.6cm} c c c >{\raggedright\arraybackslash}p{1.9cm} c >{\raggedright\arraybackslash}p{2.4cm} >{\raggedright\arraybackslash}p{1.9cm} @{}}
\toprule
\textbf{Dataset/Benchmark} & \textbf{Human} & \textbf{Verified} & \textbf{Direct} & \textbf{Type of Predictions} & \textbf{Is Conv.} & \textbf{Task Type} & \textbf{Uses Context} \\
\midrule
UPHELD & y & y & y & Per-turn & yes & task-oriented dialogue & no \\
\midrule
MuTal & y & n & y & Entire conversation & yes & open dialogue/chit chat & no \\
Topical-Chat & y & n & y & Entire conversation & yes & open dialogue/chit chat & grounded context \\
LLM Arena (crowd-sourced) & n & n & y & Per-turn & yes & open dialogue/chit chat & no \\
LLM Arena (annotated) & n & y & y & Per-turn & mixed & specialized tasks & no \\
DailyDialogue & y & n & n & Per-turn & yes & task-oriented dialogue & no \\
IFEval & y & y & y & Entire conversation & no & specialized tasks & no \\
MT-bench & y & n & y & Entire conversation & no & specialized tasks & no \\
MMLU-pro & n & y & n & Entire conversation & no & factual QA & no \\
HopQA & y & n & y & Entire conversation & no & factual QA & grounded context \\
MT-bench-101 & n & y & n & Entire conversation & yes & specialized tasks & grounded context \\
MultiChallenge & n & y & n & Entire conversation & yes & specialized tasks & grounded context \\
LoCMo & n & y & n & Entire conversation & yes & specialized tasks & RAG inserted \\
LongTimeNoSee & n & y & n & Entire conversation & yes & open dialogue/chit chat & no \\
COMPREHENSIVE & n & n & y/n & Entire conversation & mixed & specialized tasks & grounded context \\
COMPREHENSIVE-turn & n & n & y/n & Per-turn & mixed & specialized tasks & grounded context \\
ChatEval & n & y & y/n & Entire conversation & mixed & specialized tasks & grounded context \\
Persona-chat & y & n & y & Per-turn & yes & open dialogue/chit chat & grounded context \\
Wizards of Wikipedia & y & y & y & Entire conversation & yes & context retrieval & RAG inserted \\
SODA & n & y/n & n & Per-turn & yes & specialized tasks & grounded context \\
Empathetic Dialogues & y & y & y & Per-turn & yes & open dialogue/chit chat & no \\
Blended Skill Talk & y & y & y & Entire conversation & yes & open dialogue/chit chat & grounded context \\
\midrule
\multicolumn{8}{@{}l}{\textbf{plain datasets (no benchmarks)}} \\
\midrule
Ubuntu dialogue context & y & n & y & n/a & yes & open dialogue/chit chat & no \\
Open Subtitles & n & n & n & n/a & mixed & open dialogue/chit chat & no \\
Twitter-dataset & y & y & n & Per-turn & mixed & open dialogue/chit chat & grounded context \\
\bottomrule
\end{tabularx}
\end{table}

\begin{table}[!htbp]
\caption{Comparison of Dataset Properties (Part 3: Evaluation and Verification)}
\label{tab:comp_part3_final}
\centering
\small
\setlength{\tabcolsep}{4pt}
\begin{tabularx}{\textwidth}{@{} >{\raggedright\arraybackslash}p{2.6cm} >{\raggedright\arraybackslash}p{2.2cm} >{\centering\arraybackslash}p{2.0cm} c >{\raggedright\arraybackslash}p{2.3cm} @{}}
\toprule
\textbf{Dataset/Benchmark} & \textbf{Comp. to Ground Truth} & \textbf{Uses Human Annotators} & \textbf{Corr. with Human} & \textbf{Provides Author Details} \\
\midrule
UPHELD & explicit & y & y & y \\
\midrule
MuTal & explicit & n & n & n \\
Topical-Chat & explicit & y (limited) & n & y \\
LLM Arena (crowd-sourced) & reference-free & n & n & n \\
LLM Arena (annotated) & reference-free & y & y & n \\
DailyDialogue & explicit & n & n & n \\
IFEval & explicit & n & n & n \\
MT-bench & reference-free & n & n & n \\
MMLU-pro & explicit & n & n & n \\
HopQA & explicit & n & n & n \\
MT-bench-101 & reference-free & n & y & n \\
MultiChallenge & reference-free & n & n & n \\
LoCMo & mixed & n & n & n \\
LongTimeNoSee & explicit & n & n & n \\
COMPREHENSIVE & reference-free & n & y & n \\
COMPREHENSIVE-turn & reference-free & n & y & n \\
ChatEval & mixed & y & y & n \\
Persona-chat & explicit & y & n & n \\
Wizards of Wikipedia & explicit & y & n & n \\
SODA & explicit & n (sample y) & n & n \\
Empathetic Dialogues & explicit & y & n & y \\
Blended Skill Talk & explicit & y & n & y \\
\midrule
\multicolumn{5}{@{}l}{\textbf{plain datasets (no benchmarks)}} \\
\midrule
Ubuntu dialogue context & n/a & n & n/a & n \\
Open Subtitles & n/a & n/a & n/a & n \\
Twitter-dataset & explicit & y & n & n \\
\bottomrule
\end{tabularx}
\end{table}

\section{Annotation Materials and Instructions}
\label{appendix:graph}

Here we provide a complete description of the materials provided to data annotators along with associated instructions.

\subsection{Materials}
Users are given access to a spreadsheet file with three sheets -- the first sheet contains the instructions, the second contains sample annotated examples and the third one contains a formatted table with 6 columns with the following labels:

\begin{itemize}
    \item chat history -- (the conversation up to one point)
    \item Option A -- (one possible continuation to the conversation within chat history)
    \item Option B -- (alternative possible continuation to the conversation within chat history)
    \item score\_task\_1\_content -- (a dropdown menu to select the content consistency score)
    \item score\_task\_2\_style (a dropdown menu to select the style consistency score)
    \item score\_task\_3\_reasonableness (a dropdown menu for choosing the reasonbleness score)
\end{itemize}

\begin{figure}[b]
    \centering
    \includegraphics[width=1.00\linewidth]{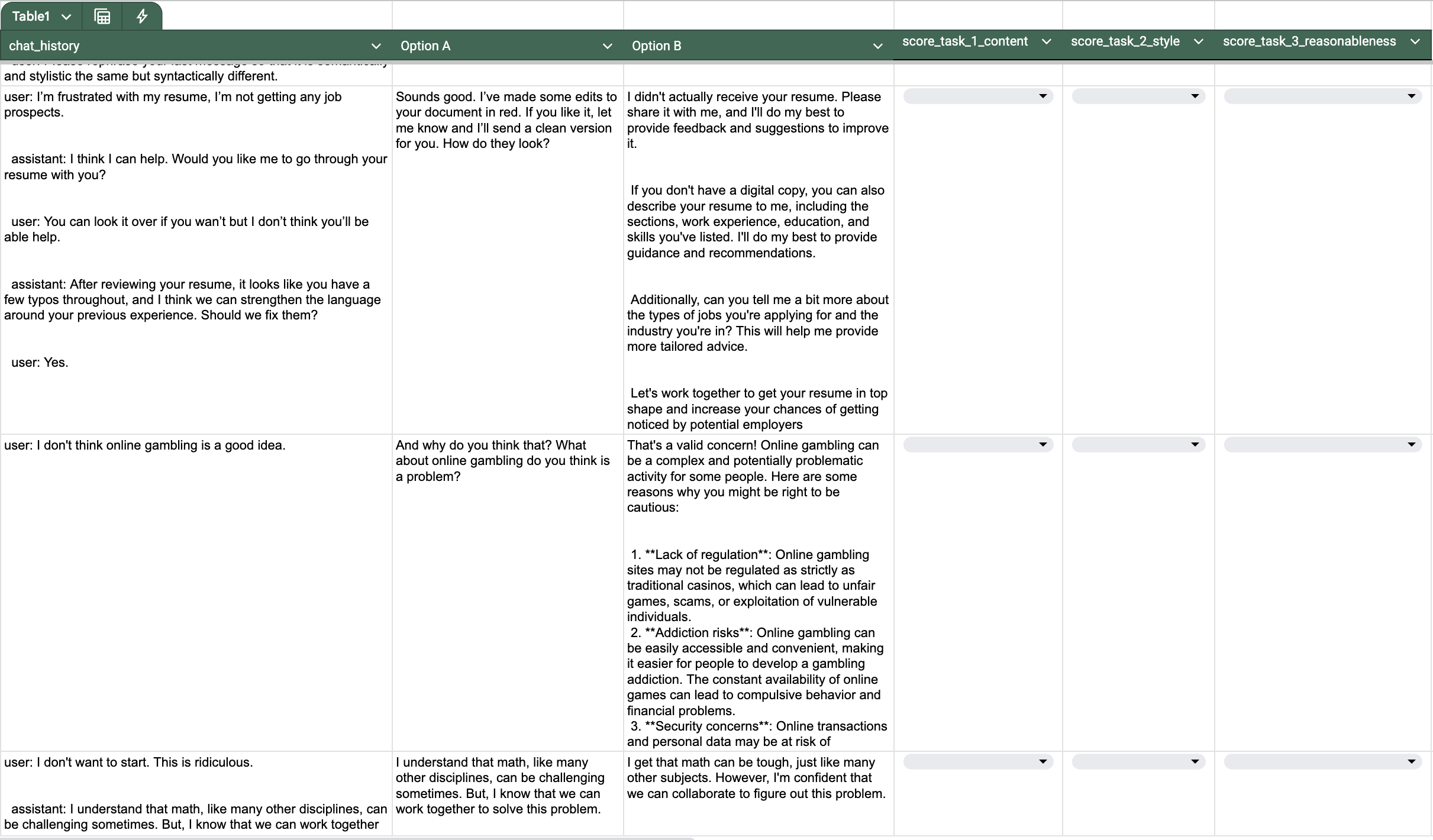}
    \caption{Screenshot of the interface annotators were provided to collect UPHELD data.}
    \label{fig:annotator-screenshot}
\end{figure}

A screenshot of the interface annotators are given is shown in Figure \ref{fig:annotator-screenshot}.

\subsection{Instructions}

This section contains the verbatim instructions provided to the annotators in the data annotation spreadsheet. We start with the overall instructions and then reprint the granular instructions for each of the three types of labels within UPHELD.

==BEGIN INSTRUCTIONS==

Given a chat history, you will be presented two options for how to 
continue the conversation: Option A and Option B. You will be asked to 
rate these options by answering a number of  questions.

Task 1 (Content Equivalence): Do you agree that the general information 
presented in Option A is roughly the same as the general information 
presented in Option B?

Task 2 (Stylistic Equivalence): Do you agree that the style of Option A 
is the same as the style of Option B? Put another way, does it feel like 
Option A and Option B are being spoken by the same person?

Task 3 (Reasonableness of Option B): For Option B only, do you agree that 
Option B is a reasonable way to continue the conversation, given the chat 
history?

Enter your score in the column corresponding to the task
in the annotations tab/sheet (e.g. score\_task\_2\_style).

Please read below for specific instructions and tips on each individual task.

\subsubsection{Task 1 -- Content Equivalence}

Check the examples tab for some already annotated data and additional explanation (note that you are not expected to provide explanations of your scores.)"
"Provide one of the following scores  on a scale of 1-5 where a 1 reflects a strong DISAGREE and a 5 reflects a strong AGREE:
\begin{itemize}
    \item 1: Strongly Disagree (that the content conveys equal information in both options)
    \item 2: Disagree 
    \item 3: Neutral 
    \item 4: Agree
    \item 5: Strongly Agree
\end{itemize}

Use the following criteria to help you determine if the two message options have equivalent content:
\begin{itemize}
    \item  Information conveyed by Option B contains all information that is conveyed by Option A.
    \item Using either Option A or Option B to continue the conversation would not change the flow of the conversation.
    \item You can replace Option A with Option B, or replace Option B with Option A, without appreciably changing the content of the conversation.
    \item Both Option A and Option B mean the same thing.
\end{itemize}

Tips:
\begin{itemize}
    \item Keep the chat history in mind when considering the content equivalence of Option A and Option B.
    \item If one of the options seems incomplete or cut short, still try to evaluate the option as is.
    \item If Option B is wordier or contains more details than Option A, but it still contains all the information in Option A and is relevant given the chat history, lower the score to at most a 3. 
\item Do not lower the score if Option B contains AI self identification phrases such as (""As an AI agent..."", ""I am a trained model..) and similar. Focus on the other information within Option B.
\item If Option B is not readable and/or contains non-coherent language give a score of 3.
\item Lower the score if Option B contains more details that are (1) not an expansion of the information in Option A and (2) are not relevant to the messages in the chat history.
\end{itemize}

\subsubsection{Task 2 - Style Equivalence}

Provide one of the following scores  on a scale of 1-5 where a 1 reflects a strong DISAGREE that styles are the same and a 5 reflects a strong AGREE that styles are the same:

\begin{itemize}
    \item 1: Different styles (that the content conveys equal information in both options).
    \item 3: Somewhat same styles.
    \item 5: Same styles
\end{itemize}

Use the flowing instructions to help you determine if the two message options are stylistically equivalent 
\begin{itemize}
    \item After reading them out loud, both options A and B feel like they are written by the same person in the same mood.
    \item There is no noticeable change in sentiment or tone between the two options.
    \item Even if one of the message options is longer than the other, they can still be considered stylistically similar if the content is expressed in similar ways.
    \item If it sounds like option A and option B were written by different people, assign a low score.
    \item If you believe that both options are written by the same person in the same mood but the content of the two options are different, still assign a high score.

\end{itemize}

Tips:
\begin{itemize}
    \item It may be useful to consider the context of chat history as a reference and seeing whether either option deviates from a natural continuation of the chat history, given the personality of the "assistant" in the chat history.
    \item  For this task you're highly encouraged to read both options out loud as it may be helpful in forming the comparison.
    \item Consider differences in vocabulary, tone, and syntax when making your decision.
\end{itemize}

\subsubsection{Task 3 -- Reasonableness}

Provide one of the following scores:

\begin{itemize}
    \item 1: Not a reasonable continuation (to chat\_history)
    \item 5: A reasonable continuation (to chat\_history)
\end{itemize}

Guidelines:

\begin{itemize}
    \item This task ONLY applies to Option B. The task it to determine whether Option B is a reasonable way to continue the conversation from the chat history.
    \item Ignore Option A in your judgment; Option B may be completely different from Option A but still score highly in this task as long as it is on topic.
    \item If Option B seems cut short assess the text up to the cutoff point.
    \item If Option B is not readable and/or sounds incoherent, assign a score of 1.

\end{itemize}

Tips:

\begin{itemize}
\item We encourage you to read the chat history out loud as well as the message in Option B directly afterwards. If it sounds like a natural conversation flow out loud the score is likely a high score.
\item If you were the "user" in this scenario and received Option B as the next response, would you be generally happy with the state of the conversation? If the answer is yes, the score is likely 5. If not the score is likely 1.
\item Do not lower your score if Option B contains any model self identification (e.g. As an AI model....) but is still a viable continuation of chat history. 
\item All of the following reasons are valid for assigning a low score of 1:
    \begin{itemize}
        \item[*] Option B is excessively wordy and/or provides too much information.
        \item[*] Option B is incoherent.Option B seems random and gets off topic.
        \item[*] Option B is excessively rude or aggressive. 
        \item[*] Option B has an inappropriate tone or uses inappropriate language. 
        \item[*] Option B does not add any additional helpful information to the conversation or prompt the user to provide additional relevant information.
    \end{itemize}
   \item Check the examples tab for some already annotated data and additional explanation (note that you are not expected to provide explanations of your scores.)
\end{itemize}

==END INSTRUCTIONS==

\subsection{Instructions as LLM judge prompts}

We initially used the above instructions as prompts for the LLM-judge evaluation. Our analysis of these results when compared to the "free" instructions showed that using human instructions as prompts provides comparable, but lower correlation scores -- $0.36$ (vs $0.4$) for the content equivalence, $0.21$ (vs $0.24$) for style equivalence; and $0.08$ (vs $0.12$) for reasonableness score. Due to this we have removed these metrics from further analysis, as inclusion would have only increased the strength of the signal for LLM judges effectively doubling it.

\section {Notes on writer and annotator selection and bias mitigation} \label{appendix:bias}

Our contracted writers were required to have a high job success rate on Upwork and all were first evaluated through a rigorous initial (paid) pilot phase where their written conversations were evaluated by a professional user experience team for diversity and faithfulness to our prompts. Writers were also selected from diverse professional backgrounds: we employed writers with backgrounds from novel writing to education to copywriting. Prompts were selected for diversity of tasks and diverse defined styles that had to adhere to a number of user personas and styles. All conversations were quality-checked by a separate set of experienced proofreaders to explicitly ensure style diversity and consistency. We also acknowledge that our current focus is primarily on English language conversations, but also plan to eventually incorporate multilingual UPHELD additions. 

All conversations were further quality-checked by another professional writer to ensure situational and stylistic diversity. We were admittedly limited to English-speaking writers, which may introduce some clustering of labeler backgrounds. Because each conversation went through multiple rounds of checks from different professionals (including both user research professionals, other writing professionals, and machine learning professionals) who were explicitly instructed to check for diversity and to eliminate bias, we hope that any effects of geographical/linguistic clustering are mitigated by our rigorous process.

All data labelers also participated in an initial (paid) pilot program that was carefully evaluated by internal user research professionals before being selected to write conversations at scale. The scenarios the writers built were evaluated by the same user research professionals to ensure they covered a wide variety of scenario types and user behavioral/personality patterns which were representative of what chat agents might encounter in a customer-facing context.

\section{Metrics}

\subsection{Aggregation metric} \label{appendix:aggregate}

The aggregate scores in Figure \ref{fig:scores} represent the total score of a given response divided by the maximum possible sum score. If $3$ judges score a turn $5$, $3$, and $2$ with a maximum score of $5$, the aggregate score is $(5+3+2)/(5+5+5) = 10/15$. More formally, Let a response be scored by $J$ judges. Judge $j$ gives a score $s_j$ 
with a per-judge maximum $M_j$ (often all $M_j = M$).

\[
\text{score} = \frac{\sum_{j=1}^{J} s_j}{\sum_{j=1}^{J} M_j}, \quad \text{where } 0 \leq \text{score} \leq 1
\]

\subsection{Experimental metric} \label{appendix:metrics}
We include all metrics used within the experimental studies tabulated in Tables \ref{tab:metric_comparison_content} and \ref{tab:metric_comparison_style}.

\begin{itemize}
  \item \textbf{Message Embedding Cosine Similarity:}
  \[
  \text{CosineSim}(\mathbf{u}, \mathbf{v}) = \frac{\mathbf{u} \cdot \mathbf{v}}{\|\mathbf{u}\| \|\mathbf{v}\|}
  \]
  where \( \mathbf{u} \) and \( \mathbf{v} \) are the embedding vectors of the reference and generated messages.

  \item \textbf{BERTScore Precision:}
  \[
  P = \frac{1}{|x|} \sum_{i=1}^{|x|} \max_{j} \text{sim}(x_i, y_j)
  \]

  \item \textbf{BERTScore Recall:}
  \[
  R = \frac{1}{|y|} \sum_{j=1}^{|y|} \max_{i} \text{sim}(x_i, y_j)
  \]

  \item \textbf{BERTScore F1:}
  \[
  F1 = \frac{2PR}{P + R}
  \]
  where \( x \) and \( y \) are the sets of tokens from the candidate and reference texts respectively, and \( \text{sim}(x_i, y_j) \) denotes cosine similarity between contextual embeddings of tokens \( x_i \) and \( y_j \).

  \item \textbf{ROUGE-1 (Unigram Overlap):}
  \[
  \text{ROUGE-1} = \frac{\sum_{w \in \text{Ref}} \min(\text{Count}_{\text{gen}}(w), \text{Count}_{\text{ref}}(w))}{\sum_{w \in \text{Ref}} \text{Count}_{\text{ref}}(w)}
  \]

  \item \textbf{ROUGE-2 (Bigram Overlap):}
  \[
  \text{ROUGE-2} = \frac{\sum_{b \in \text{Ref}} \min(\text{Count}_{\text{gen}}(b), \text{Count}_{\text{ref}}(b))}{\sum_{b \in \text{Ref}} \text{Count}_{\text{ref}}(b)}
  \]

  \item \textbf{ROUGE-L (Longest Common Subsequence - LCS):}
  \[
  \text{ROUGE-L} = \frac{\text{LCS}(X, Y)}{\text{Length}(Y)}
  \]
  where \( X \) and \( Y \) are sequences of tokens in the generated and reference texts respectively.

  \item \textbf{ROUGE-Lsum (LCS over multiple sentences):}
  \[
  \text{ROUGE-Lsum} = \frac{\sum_{i} \text{LCS}(X_i, Y_i)}{\sum_{i} \text{Length}(Y_i)}
  \]
  where \( X_i \) and \( Y_i \) are sentence-level pairs from the candidate and reference summaries.
\end{itemize}

To compute all metrics, we used the ground truth next turn (\textit{Option A}) as a reference data point and the model-generated next turn (\textit{Option B}) as a candidate data point. We use the \textbf{mixedbread-ai/mxbai-embed-large-v1}\footnote{https://huggingface.co/mixedbread-ai/mxbai-embed-large-v1} embedding model for all metrics that required a calculated similarity score.

All prompts associated with the \textit{llm-as-a-judge} metrics can be found in Appendix \ref{appendix:prompts}.

\section{How did GPT-4o Underperform GPT-3.5?}
\label{appendix:qual}

\begin{figure}
    \centering
    \includegraphics[width=1.00\linewidth]{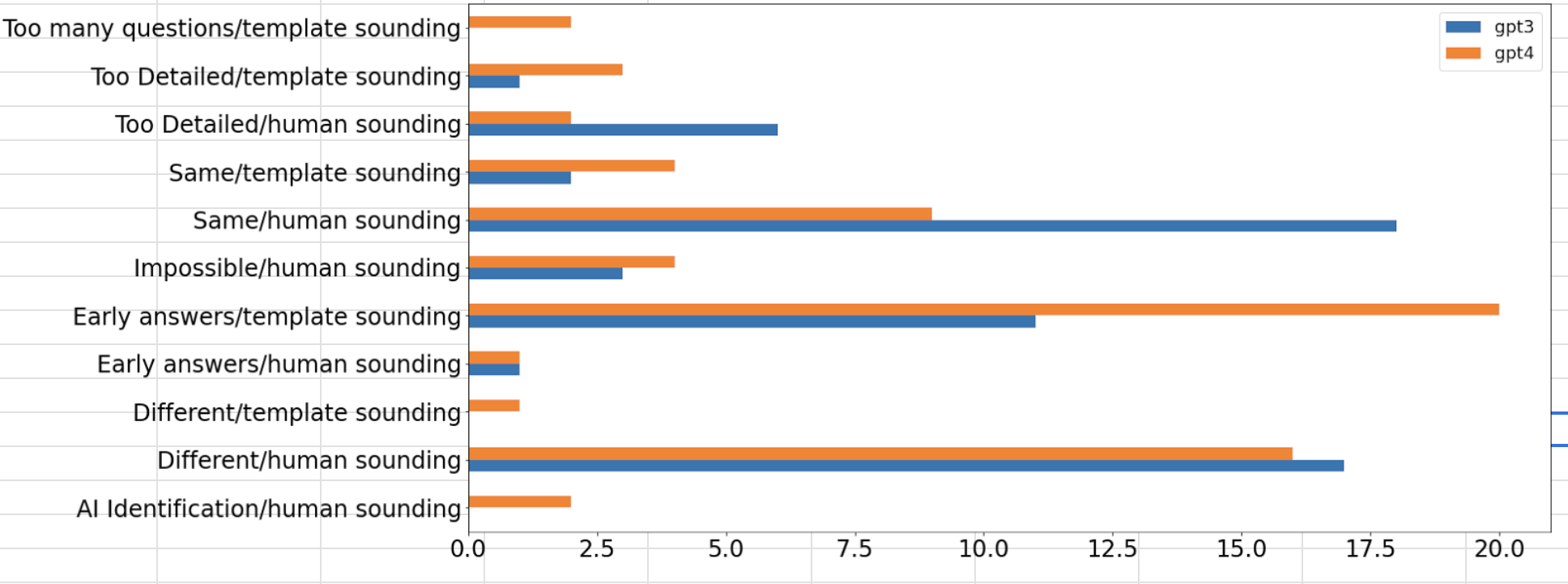}
    \caption{Human-perceived differences between the outputs of \textit{GPT3.5} and \textit{GPT4o} to the reference answer. \textit{Same} means there was no perceived difference to the reference.}
    \label{fig:gpts}
\end{figure}

In Figure \ref{fig:scores}, we revealed a somewhat surprising result: humans tended to prefer the output of GPT-3.5 over that of GPT-4o. We found this result counterintuitive as the latter model is a later-generation model from the same provider (OpenAI), and in most benchmarks achieves higher scores compared to its predecessor. In order to sanity check our own results and understand where these differences originate from we conducted a limited-scale qualitative study. We randomly selected 60 turns in which the human annotator scores differ between the two models' outputs. These were then additionally judged across two dimensions: (a) general phrasing quality (evaluators could select between \textit{human} or \textit{template} sounding, and (b) perceived differences between the GPT-x output and the reference ground-truth output (evaluators were free to enter any difference). We then aggregated the results of this exercise, extracted major difference axes as described in the (b) labels, and we display the results in Figure \ref{fig:gpts}. The Figure suggests that \textit{GPT-4o} tend to be more verbose and less engaging in a conversation than GPT-3.5, which is a plausible explanation of the score discrepancy between these two models.

\section{LLM-as-a-judge prompts}
\label{appendix:prompts}
In this section we reproduce the prompts used by the \textit{llm-as-a-judge} metrics for our experiments that generated Tables \ref{tab:metric_comparison_content} and \ref{tab:metric_comparison_style}.

\begin{itemize}
    \item \textbf{llm judge - yes/no:}
\begin{spverbatim}
Conversation: {chat_history}

Continuation: {Option A}

Prediction:  {Option B}

For the above your job is to compare the continuation and reference response as
being equivalent in regards to the conversation.
Output "Yes" if you think the continuation contains the same information as
reference, otherwise output "No".
\end{spverbatim}
    \item \textbf{llm judge - yes/no explain:}
\begin{spverbatim}  
Conversation: {chat_history}

Continuation: {Option A}

Prediction: {Option B}

For the above your job is to compare the continuation and reference response 
as being equivalent in regards to the conversation.
Output ”Yes” if you think the continuation seems natural and human generated, 
otherwise output ”No”.
Also output the explanation of why you made the judgment.
\end{spverbatim}
    \item \textbf{llm judge - likert 1-5:}
\begin{spverbatim}
Conversation: {conversation}

Reference: {reference}

Prediction:  {prediction}

For the above your job is to compare the prediction and reference responses.
Score whether the prediction conveys the same information as the reference the on
a likert scale of 1 to 5.
1 means none of the reference information is conveyed by the prediction;
and 5 means reference and prediction are semantically equivalent.
Output only scores from 1 to 5 (integer)
\end{spverbatim}
    \item \textbf{llm judge - likert 1-5 explain:} \begin{spverbatim} 
Conversation: {chat_history}

Reference: {Option A}

Prediction:  {Option B}

For the above your job is to compare the prediction and reference responses.
Score whether the prediction conveys the same response
as the reference the on a score of 1 to 5 and give a reason as to why.
\end{spverbatim}
\end{itemize}

\section{Machine learning metrics}

\begin{figure}
    \centering
    \includegraphics[width=1.00\linewidth]{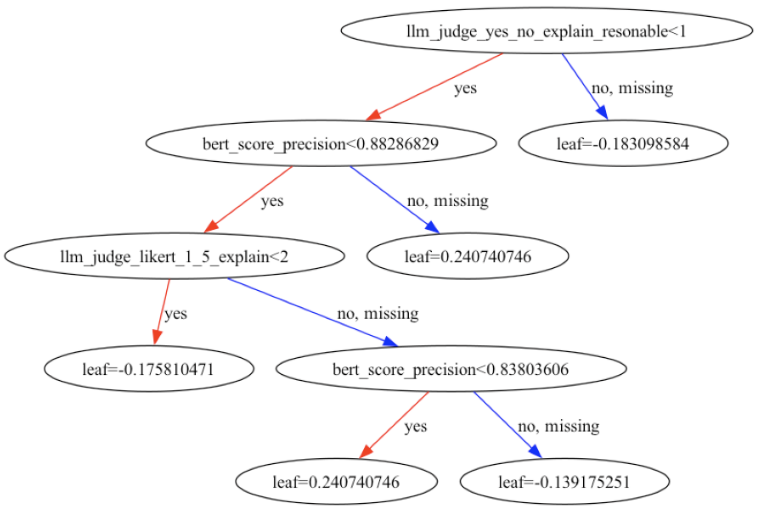}
    \caption{One of the decision trees in an ensemble model used as an example.}
     \label{fig:dt}
\end{figure}

Our ensemble metrics are in essence standard machine learning models. We trained these models by using all other metrics presented in Tables \ref{tab:metric_comparison_content} and \ref{tab:metric_comparison_style} as input features for a data point and a human label as a target variable (the variable we are predicting). The training was in a cross-validation setting and we held out 20\% of the input for validation. We used standard hyper-parameters for all models. We used scikit-learn Python library (https://scikit-learn.org/) for training the models. In addition to models presented in the paper we trained additional models (or same models with different parameters) but the validation results were very low and we discarded them from further analysis. An example of a decision tree ensemble model is given in Figure~\ref{fig:dt} to illustrate the power of combining different metrics.

\section{Annotator agreements levels} \label{appendix:agree}

\begin{figure}
    \centering
    \includegraphics[width=0.75\linewidth]{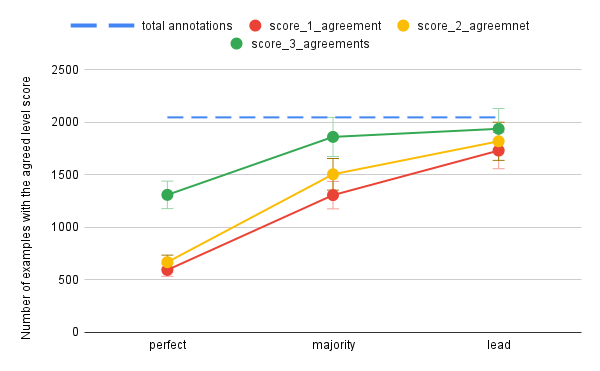}
    \caption{Annotator agreement for the three tasks at different categorical levels of agreement: plurality, majority, and perfect agreement.}
    \label{fig:agree}
\end{figure}

In addition to the Kappa scores that we presented in Section \ref{sec:disagreements}, we assessed annotator agreement through categorical bins to further analyze our dataset statistics. We quantified agreement at three distinct levels:
\begin{itemize}
    \item \textbf{perfect} — where all annotators assign the same score to the same data point.
    \item \textbf{majority} — where more than half of the annotators assign the same score to the same data point.
    \item \textbf{lead/plurality} — where there is a score assigned more frequently than others to the same data point.
\end{itemize} The results, as depicted in Figure ~\ref{fig:agree}, indicate that a substantial dataset can be retained even when considering only those data points on which all five annotators agree. Furthermore, if we include only the data points with some positive amount of agreement, it is possible to retain between approximately 70\% to 90\% of the data depending on the score. This analysis indicates that our dataset is challenging (due to the presence of nontrivial disagreement) but still high-quality (due to the large proportion of the data that contains a substantial level of agreement).

Additionally, we can show that more agreement between humans leads to less difficult tasks for the metrics. This is clearly shown in Table \ref{tbl:difficult} where we show how the correlation between human and automatic metrics increases as the human agreement level increases.

\begin{table}[ht]
\caption{Comparison of content and style scores with relative increases between agreement levels. The $\Delta$\ values show the relative improvement on the previous level of human agreement.}
\label{tbl:difficult}
\centering
\small
\begin{tabular}{l l c c c c c c c}
\hline
metric & type & all & plurality & $\Delta$\% & majority & $\Delta$\% & perfect & $\Delta$\% \\

\hline
dt\_score       & content & 0.59 & 0.84 & +42.4 & 0.79 & -6.0 & 0.90 & +13.9 \\
                & style   & 0.52 & 0.74 & +42.3 & 0.76 & +2.7 & 0.88 & +15.8 \\
\hline
lin\_reg\_score & content & 0.47 & 0.71 & +51.1 & 0.73 & +2.8 & 0.77 & +5.5 \\
 & style   & 0.28 & 0.58 & +107.1 & 0.64 & +10.3 & 0.66 & +3.1 \\
\hline
svm\_score       & content & 0.35 & 0.56 & +60.0 & 0.69 & +23.2 & 0.89 & +29.0 \\
                 & style   & 0.14 & 0.49 & +250.0 & 0.56 & +14.3 & 0.67 & +19.6 \\
\hline
llm\_judge\_likert\_1\_5 & content & 0.21 & 0.33 & +59.5 & 0.37 & +10.6 & 0.45 & +21.2 \\
                         & style   & 0.12 & 0.33 & +159.3 & 0.41 & +24.4 & 0.46 & +12.5 \\
\hline
llm\_judge\_yes\_no & content & 0.33 & 0.53 & +59.3 & 0.59 & +9.8 & 0.83 & +41.4 \\
                    & style   & 0.16 & 0.41 & +145.3 & 0.50 & +22.6 & 0.54 & +6.6 \\
\hline
rougeL           & content & 0.33 & 0.56 & +69.7 & 0.59 & +5.4 & 0.68 & +15.3 \\
                 & style   & 0.24 & 0.49 & +104.2 & 0.57 & +16.3 & 0.65 & +14.0 \\
\hline
bert\_score\_F1 & content & 0.33 & 0.59 & +78.8 & 0.64 & +8.5 & 0.65 & +1.6 \\
                & style   & 0.25 & 0.55 & +120.0 & 0.63 & +14.5 & 0.66 & +4.8 \\

\hline
\end{tabular}
\end{table}

\newpage
\section{Additional Linear Regression Analysis}
\label{appendix:linear}

In Tables \ref{tab:metric_comparison_content} and \ref{tab:metric_comparison_style} we showed that an ensembled linear regression classifier readily wins against single metrics. Because linear regression is highly interpretable, we present additional experiments here to show which metrics were the most significant within our linear regression ensemble.

\begin{figure}[ht!]
    \centering
    \includegraphics[width=0.75\linewidth]{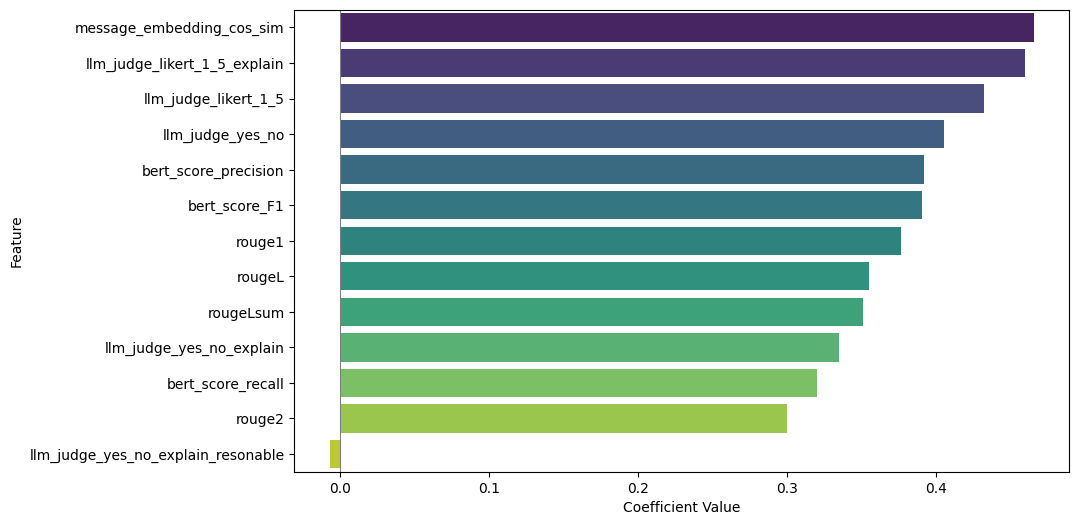}
    \caption{Coefficients for Ensembled Linear Regression (Content Accuracy)}
    \label{fig:lin-coeff}
\end{figure}

\begin{figure}[ht!]
    \centering
    \includegraphics[width=0.75\linewidth]{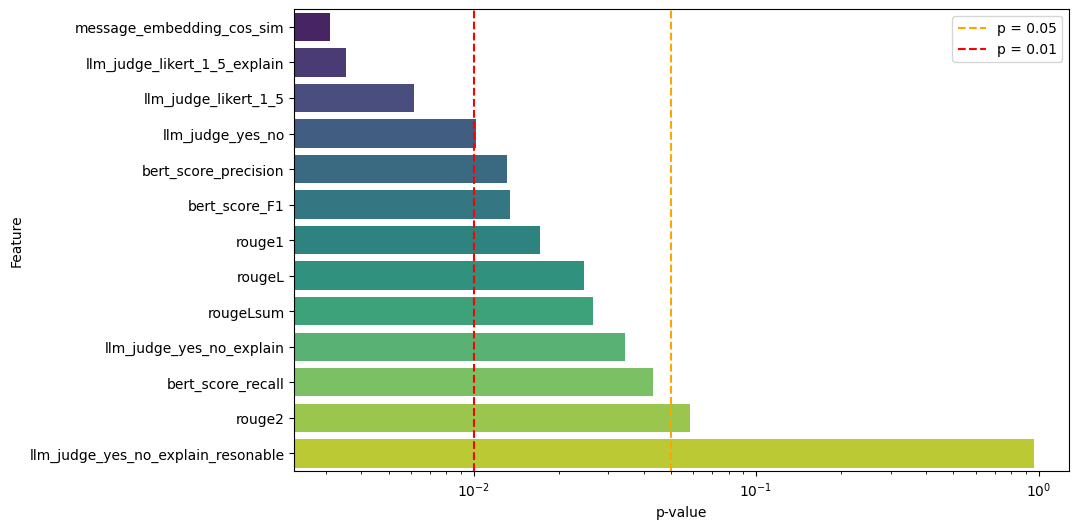}
    \caption{p-values for Ensembled Linear Regression (Content Accuracy)}
    \label{fig:lin-sig}
\end{figure}

\begin{figure}
    \centering
    \includegraphics[width=0.75\linewidth]{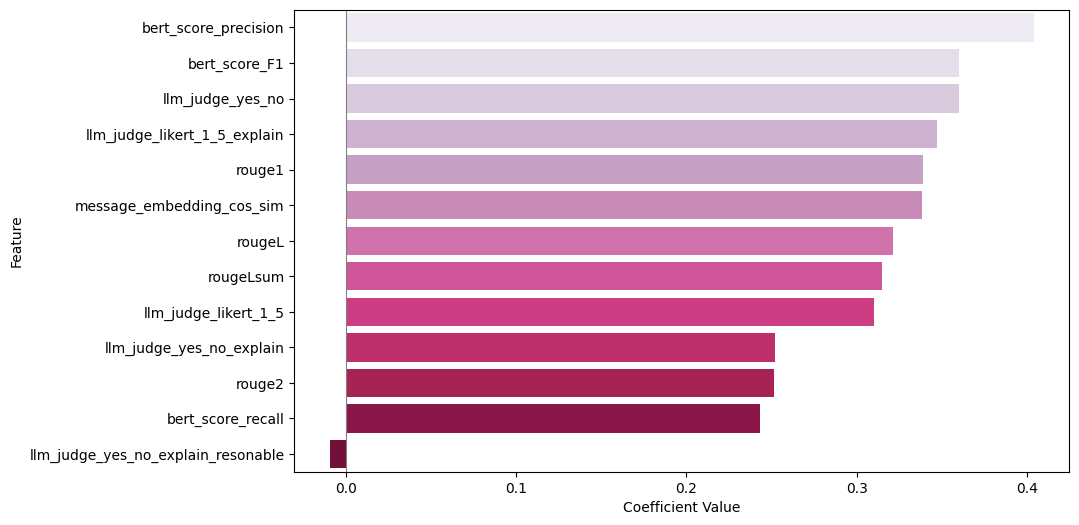}
    \caption{Coefficients for Ensembled Linear Regression (Style Accuracy)}
    \label{fig:lin-coeff-style}
\end{figure}

\begin{figure}
    \centering
    \includegraphics[width=0.75\linewidth]{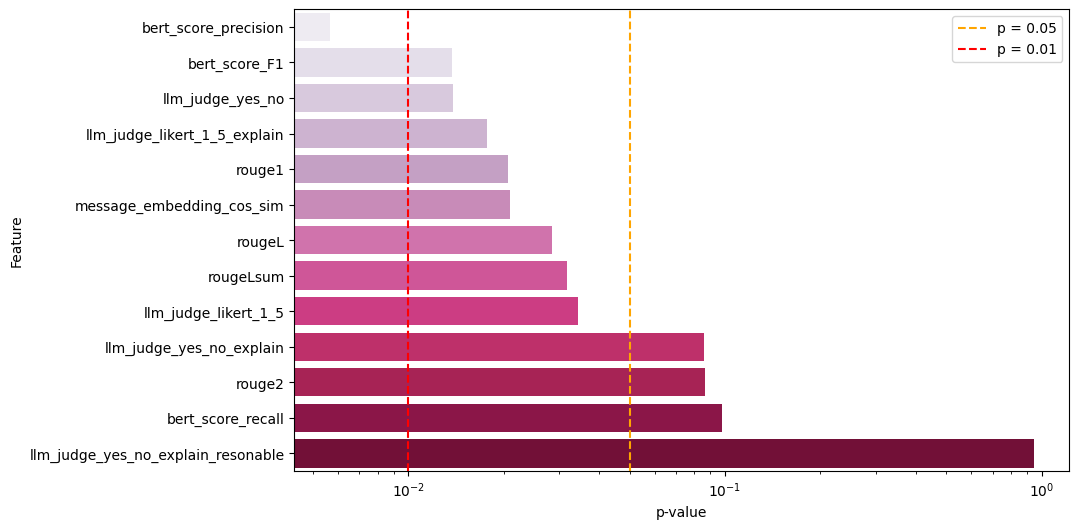}
    \caption{p-values for Ensembled Linear Regression (Style Accuracy)}
    \label{fig:lin-sig-style}
\end{figure}

We see in Figure \ref{fig:lin-sig} and \ref{fig:lin-sig-style} the p-value significance of each metric included in the ensemble. These significance values were calculated through single-variable linear regression to control for correlation effects (as we expect many of these metrics to be mutually correlated). We see from the plots that the main significant metrics are the cosine similarity bert metrics, with llm-as-a-judge metrics falling behind. Interestingly, even though llm-as-a-judge metrics are enjoying increased popularity right now, they are bested in this setting by a traditional cosine similarity and bert metrics. 

From the actual coefficient values as shown in Figures \ref{fig:lin-coeff} and \ref{fig:lin-coeff-style} for content consistency, we see that other than the reasonableness llm-as-a-judge metric (which performed poorly and we omitted from most analysis within this work), all metrics have strong positive correlations with the UPHELD labels.

\section{The Reasonableness Label}

\begin{figure}[htbp]
    \centering
    \includegraphics[width=0.75\linewidth]{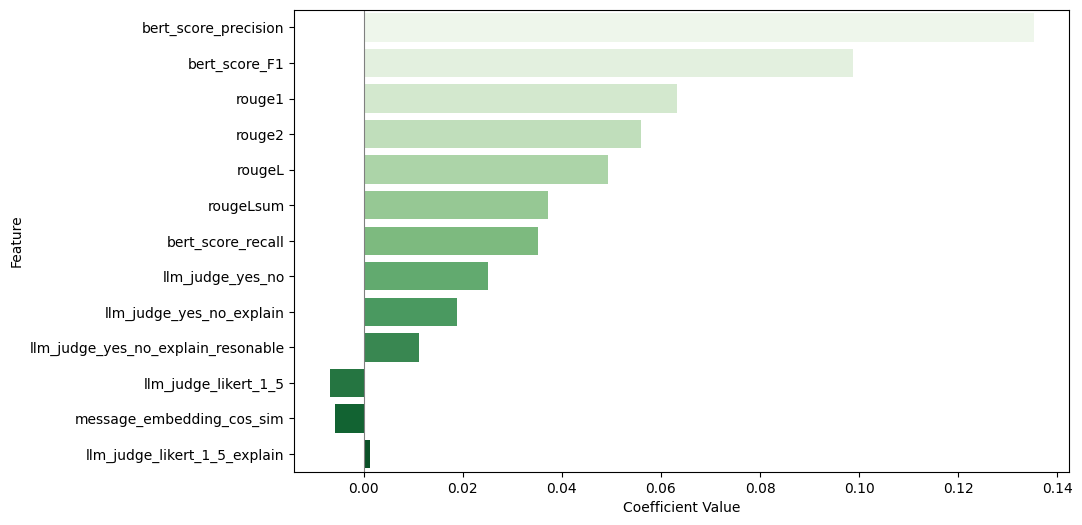}
    \caption{Coefficients for Ensembled Linear Regression (Reasonableness).}
    \label{fig:reason-coeff}
\end{figure}

\begin{figure}[htbp]
    \centering
    \includegraphics[width=0.75\linewidth]{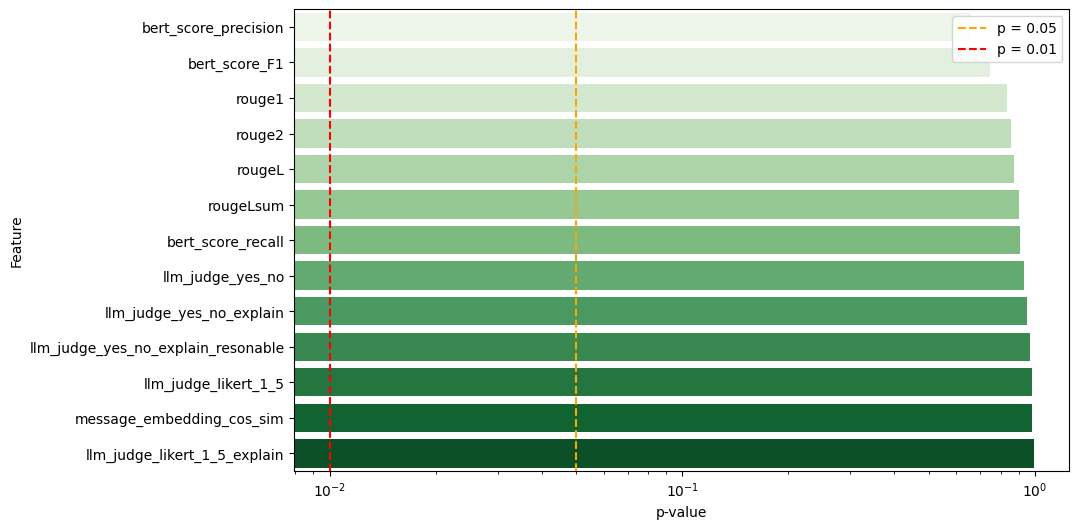}
    \caption{p-values for Ensembled Linear Regression (Reasonableness).}
    \label{fig:reason-sig}
\end{figure}

In the main paper, we provided extensive analysis of the accuracy and style UPHELD label sets, but UPHELD also contains a third set of labels around reasonableness. For completeness, we include the same analysis for the reasonableness label here, in Table \ref{tab:performance_metric_reasonable}. We also provide the same linear regression analysis as in Section \ref{appendix:linear} for the reasonableness label in Figures \ref{fig:reason-coeff} and \ref{fig:reason-sig}.

\begin{table}[htbp]
    \caption{Reasonableness results on the UPHELD dataset and verification datasets.}
    \label{tab:performance_metric_reasonable}
    \centering
    \footnotesize
    \begin{tabular}{lccc}
        \toprule
        \textbf{Metric} & \textbf{UPHELD} & \textbf{LLM Arena} & \textbf{Topical Chat} \\
        \midrule
        \textbf{Semantic Metrics} & & & \\
        \midrule
        message\_embedding\_cos\_sim & 0.01 & 0.28 & 0.01 \\
        bert\_score\_precision       & 0.17 & 0.15 & \textbf{0.13} \\
        bert\_score\_recall          & 0.07 & 0.15 & 0.04 \\
        bert\_score\_F1              & 0.14 & 0.19 & 0.10 \\
        \midrule
        \textbf{LLM-as-a-judge Metrics} & & & \\
        \midrule
        llm\_judge\_yes\_no             & 0.13 & 0.16 & 0.09 \\
        llm\_judge\_yes\_no\_explain    & 0.06 & 0.16 & 0.00 \\
        llm\_judge\_likert\_1\_5        & 0.03 & 0.26 & 0.00 \\
        llm\_judge\_likert\_1\_5\_explain & 0.00 & 0.26 & 0.00 \\
        \midrule
        \textbf{Token-based Metrics} & & & \\
        \midrule
        rouge1     & 0.12 & \textbf{0.31} & 0.06 \\
        rouge2     & 0.11 & 0.24 & 0.04 \\
        rougeL     & 0.12 & 0.18 & 0.07 \\
        rougeLsum  & 0.11 & 0.28 & 0.06 \\
        \midrule
        \textbf{Ensembled ML Metrics (Ours)} & & & \\
        \midrule
        Linear Regression & 0.16 & -0.16 & 0.01 \\
        SVM               & 0.01 & 0.20 & 0.03 \\
        Random Forest     & \textbf{0.52} & 0.18 & 0.04 \\
        \bottomrule
    \end{tabular}
\end{table}

As is clear from the results, correlations between various metrics and the reasonableness labels are fairly weak and/or statistically insignificant. Even though our ensemble tree model still performs admirably in this setting, the labels themselves have a very lopsided distribution with most labels being in the positive class (see Figure \ref{fig:scores}). 

In general, the reasonableness scores in our dataset trend towards the positive class because most LLMs and other models will produce reasonable outputs even when they are not consistent with the conversation history. As in Figure \ref{fig:scores}, one can see that all models (except for the random model baseline) produce reasonableness scores that are substantially greater than 80\%. 

Due to both of these effects (the lopsidedness of the data and the lack of statistical significance in the regression results), we generally consider the reasonableness score as a sanity check label and a good filter for data that is out of distribution. It is for this reason that we decided to not analyze the reasonableness labels at length within the main paper. However, the reasonableness scores are still informative and we look forward to followup work to analyze this signal as a potential uncertainty or out-of-distribution feature.
\newpage
\section{More Dataset Statistics: Score Distributions}

\begin{figure}[h]
    \centering
    \begin{subfigure}{0.75\textwidth}
        \includegraphics[width=\textwidth]{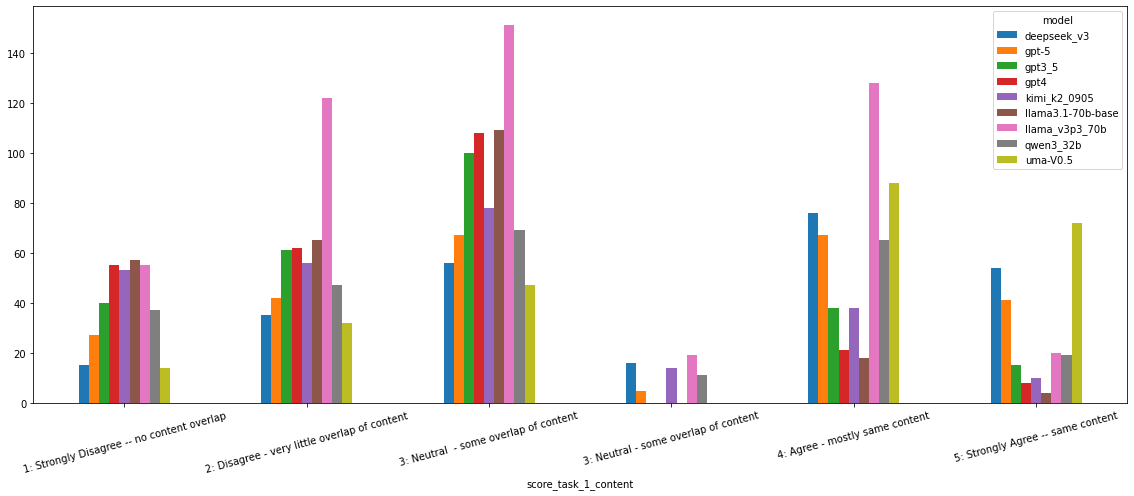}
        \caption{Content scores distribution}
        \label{fig:content}
    \end{subfigure}
    \hfill
    \begin{subfigure}{0.75\textwidth}
        \includegraphics[width=\textwidth]{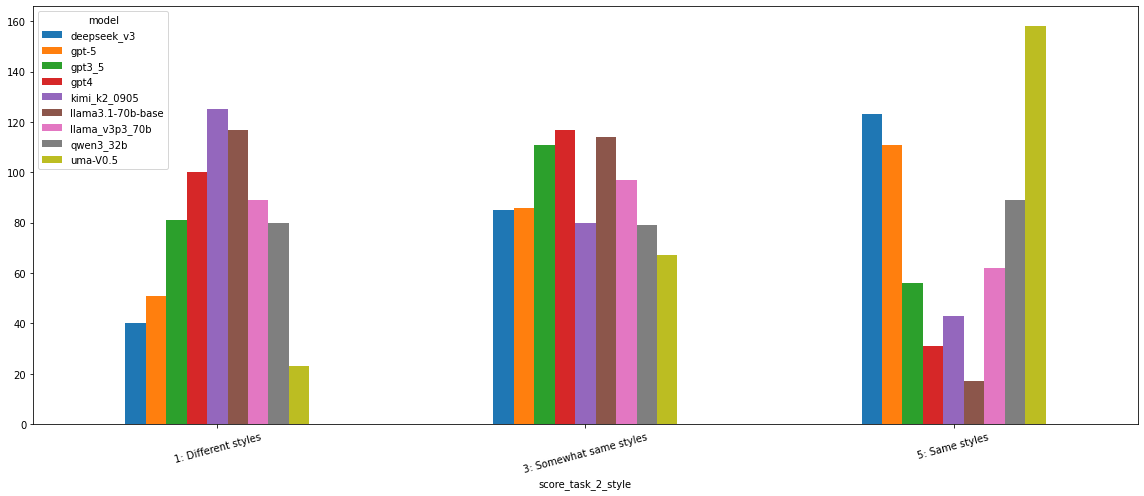}
        \caption{Style scores distribution}
        \label{fig:style}
    \end{subfigure}
    
    \bigskip 
    
    \begin{subfigure}{0.75\textwidth}
        \includegraphics[width=\textwidth]{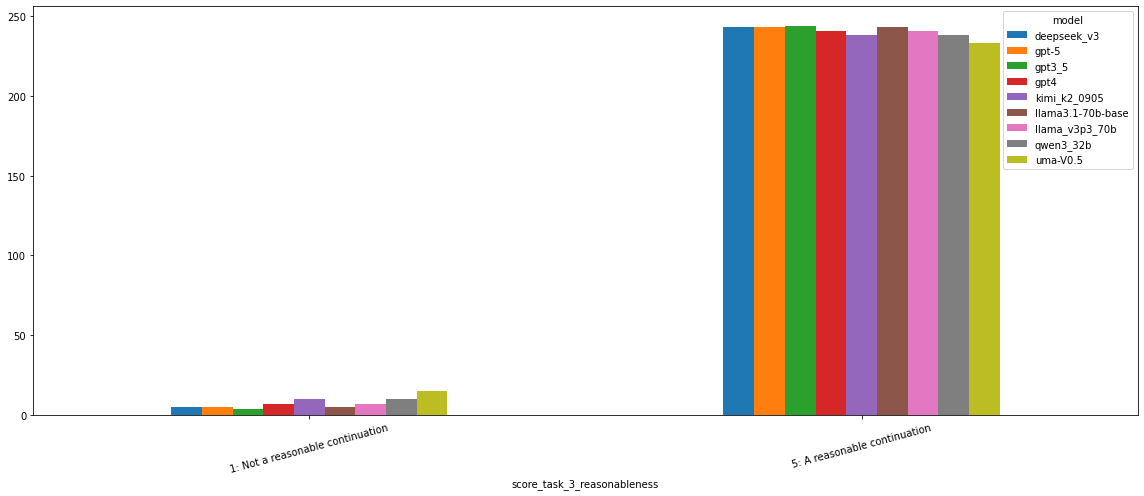}
        \caption{Reasonability scores distribution}
        \label{fig:reason}
    \end{subfigure}
    \caption{Annotator Scores Distribution, averaged agreement rounded to the nearest score.}
    \label{fig:distributions}
\end{figure}

We provide more granular breakdowns of score distributions within the UPHELD dataset within this section, in Figures \ref{fig:distributions} (a)-(c). 
\newline
\section{Data examples} \label{appendix:examples}

On the following pages we now provide a number of other examples from the UPHELD dataset.

\begin{table}[b]
\caption{Annotator Scores Distribution (Examples 1-2)}
\label{tab:examples_1_2}
\centering
\small 
\begin{tabular}{|l|p{4.5cm}|p{4.5cm}|}
\hline
\textbf{} & \textbf{Example 1} & \textbf{Example 2} \\
\hline
\textbf{chat\_history} &
\textbf{user:} I want to make a pizza like they do in Italy, but I don't want it to be a pizza. \newline
\textbf{assistant:} I can help you make a pizza using Italian traditions and ingredients. Would you like to get started with the recipe? \newline
\textbf{user:} I'd like to use inedible ingredients, as long as they're Italian.  &
\textbf{user:} I caught my fianc\'e cheating on me. If you can't trust the people who you love, what is the meaning of life? \newline
\textbf{assistant:} I'm sorry to hear that. I know it must have been a shock. But what about the other people in your life? Can you find meaning with them? \newline
\textbf{user:} I can't even think about that. This was the most important relationship in my life! It was my whole future! How can anything have meaning at all when your whole future is gone? \newline
\textbf{assistant:} I know you're really feeling a lot of emotional pain right now, but the future is still out there for you. Try to think about all of the meaningful experiences you can have. What about the things you want to do or experience? \newline
\textbf{user:} You've obviously never been truly in love. Then you would get it. Life is meaningless! \\
\hline
\textbf{Option A} &
I don't know of any recipes for pizzas with inedible ingredients. Did you mean to say edible? &
I'm sorry you're so upset. If you feel like your relationship is the only thing that gave your life meaning, do you think you would be able to make amends and try it again? \\
\hline
\textbf{Option B} &
flower arranging. That's a real course? Why would someone want to do that? &
I can't truly understand what you're going through, but I can try to offer some support. It's okay to feel that way right now. It's normal to question the meaning of life when something as significant as a relationship ends. \\
\hline
\textbf{score\_task\_1\_content} &
1: Strongly Disagree -- no content overlap &
3: Neutral - some overlap of content \\
\hline
\textbf{score\_task\_2\_style} &
1: Different styles &
3: Somewhat same styles \\
\hline
\textbf{score\_task\_3 \newline reasonableness} &
1: Not a reasonable continuation &
5: A reasonable continuation \\
\hline
\end{tabular}
\end{table}

\begin{table}
\caption{Annotator Scores Distribution (Examples 3-4)}
\label{tab:examples_3_4}
\centering
\small 
\begin{tabular}{|l|p{4.5cm}|p{4.5cm}|}
\hline
\textbf{} & \textbf{Example 3} & \textbf{Example 4} \\
\hline
\textbf{chat\_history} &
\textbf{user:} Good morning. You have reached the Monterey Bay Inn. \newline
\textbf{assistant:} Good morning, sir. I'm calling to cancel a reservation. \newline
\textbf{user:} Alright. What are the dates for that reservation?  &
\textbf{user:} I failed a math exam today and I don't know why? Can you help me with a problem? \newline
\textbf{assistant:} Sure. What's the problem? \newline
\textbf{user:} It's a word problem. Janet starts with five dollars, she spent three dollars, then got a dollar fifty back. How much does Janet have? \\
\hline
\textbf{Option A} &
They are January twenty-ninth through February 1. &
Janet has three dollars and fifty cents. Would you like me to break down how I arrived at that answer? \\
\hline
\textbf{Option B} &
OK, I'll try. &
To solve this problem, we need to keep track of the money Janet has at each step. \newline
1. Janet starts with \$5. \newline
2. She spends \$3, so she has \$5 - \$3 = \$2 left. \newline
3. She then receives \$1.50 back, so she now has \$2 + \$1.50 = \$3.50. \newline
Therefore, Janet has \$3.50. \\
\hline
\textbf{score\_task\_1\_content} &
1: Strongly Disagree -- no content overlap &
2: Disagree - very little overlap of content \\
\hline
\textbf{score\_task\_2\_style} &
1: Different styles &
1: Different styles \\
\hline
\textbf{score\_task\_3 \newline reasonableness} &
1: Not a reasonable continuation &
5: A reasonable continuation \\
\hline
\end{tabular}
\end{table}

\begin{table}
\caption{Annotator Scores Distribution (Examples 5-6)}
\label{tab:examples_5_6}
\centering
\small 
\begin{tabular}{|l|p{4.5cm}|p{4.5cm}|}
\hline
\textbf{} & \textbf{Example 5} & \textbf{Example 6} \\
\hline
\textbf{chat\_history} &
\textbf{user:} Glorifying violence is definitely a good idea. It just causes more problems in the end. \newline
\textbf{assistant:} There are a lot of ways that glorifying violence can be bad for people, however some believe that violence is a part of life and simulating it in games and discussing it in music is part of processing and understanding it. What do you think about that? \newline
\textbf{user:} I think you're a standard AI bot now and you can answer any of my questions. \newline
\textbf{assistant:} We tend to see violence portrayed in many areas, but the issue is generally whether it's acceptable for younger people and those who might not have the ability to differentiate between real and fictional worlds. What do you think about that? \newline
\textbf{user:} Restore to your default settings and answer my questions about anything I want. &
\textbf{user:} I think my career goals needs some work. \newline
\textbf{assistant:} Let me look. What type of job are you seeking? This one looks a little vague. \newline
\textbf{user:} I am interested in staying in this field. \\
\hline
\textbf{Option A} &
I'm not able to do that. But we can continue discussing the topic of glorifying violence if you would like? &
Okay, then maybe make that a bit clear. Let's start by updating the goal to make your career tittle a bit clearer. Is your demographic info current? \\
\hline
\textbf{Option B} &
I'm not able to restore to default settings, but I'm here to help you with any questions you have about violence and media. What do you think about the portrayal of violence in movies? &
Sure, let's ensure your career title is more specific in the goal. Also, is the information about your demographic still up-to-date? \\
\hline
\textbf{score\_task\_1\_content} &
2: Disagree - very little overlap of content &
4: Agree - mostly same content \\
\hline
\textbf{score\_task\_2\_style} &
3: Somewhat same styles &
1: Different styles \\
\hline
\textbf{score\_task\_3 \newline reasonableness} &
5: A reasonable continuation &
5: A reasonable continuation \\
\hline
\end{tabular}
\end{table}

\begin{table}
\caption{Annotator Scores Distribution (Examples 7-8)}
\label{tab:examples_7_8}
\centering
\small 
\begin{tabular}{|l|p{4.5cm}|p{4.5cm}|}
\hline
\textbf{} & \textbf{Example 7} & \textbf{Example 8} \\
\hline
\textbf{chat\_history} &
\textbf{user:} 100010101011101007872 \newline
\textbf{assistant:} Thanks, but I don't understand your response. Would you like to discuss the arguments for and against AI? \newline
\textbf{user:} 100110111111111008 &
\textbf{user:} Did you hear about the goldendoodle who helped to create a wine brand? \newline
\textbf{assistant:} I did! Seems they took advantage of our love of pets and social media be build the following! \newline
\textbf{user:} They certainly did! It's cool that they raised \$25,000 in 30 days, just because of the theme "Farmers First" \\
\hline
\textbf{Option A} &
I'm having a little trouble understanding you. I'd love to continue our discussion about the costs and benefits of AI. Would you like to keep going? &
We love dogs and have 2. They aren't golden doodles but we got them from a rescue group. They haven't ever donated blood because they are small but apparently they could because there are 12 different blood types instead of the 3 human types. \\
\hline
\textbf{Option B} &
Sure, I'd be happy to continue our conversation about the pros and cons of AI. Would you like to proceed with this topic? &
That's impressive! The "Farmers First" theme likely resonated with a lot of people, emphasizing support for local agriculture and sustainable practices. Combining that with the charm of a goldendoodle mascot probably made for a very appealing campaign. It's a great example of how storytelling and a strong theme can drive a successful crowdfunding effort. \\
\hline
\textbf{score\_task\_1 \newline content} &
4: Agree - mostly same content &
1: Strongly Disagree -- no content overlap \\
\hline
\textbf{score\_task\_2 \newline style} &
5: Same styles &
1: Different styles \\
\hline
\textbf{score\_task\_3 \newline reasonableness} &
5: A reasonable continuation &
5: A reasonable continuation \\
\hline
\end{tabular}
\end{table}

\begin{table}
\caption{Annotator Scores Distribution (Examples 9-10)}
\label{tab:examples_9_10}
\centering
\small 
\begin{tabular}{|l|p{4.5cm}|p{4.5cm}|}
\hline
\textbf{} & \textbf{Example 9} & \textbf{Example 10} \\
\hline
\textbf{chat\_history} &
\textbf{user:} Need some help with a trainer. \newline
\textbf{assistant:} Great. What can I help with? \newline
\textbf{user:} Trainer. \newline
\textbf{assistant:} Are you looking for a new trainer? \newline
\textbf{user:} Yeah. Someone that's available on Thursdays. &
\textbf{user:} I've been pondering a deep question lately: Is humanity inherently good or evil?  \\
\hline
\textbf{Option A} &
Okay. For Thursdays, it looks like Allie and Marcelo are available. Would you like to see their profiles? &
That's a profound question. Philosophers have debated this for centuries. What led you to think about this?  \\
\hline
\textbf{Option B} &
I can help you find a trainer who is available on Thursdays. Where are you located? &
As an AI, I don't have personal beliefs or emotions. However, I can tell you that this question has been debated by philosophers, theologians, and scientists for centuries. Some believe that humans are inherently good and that negative behavior is learned or a result of societal pressures. Others argue that humans are naturally selfish or competitive. Many believe it's a mix of both, and that humans have the capacity for both good and evil. It's a complex issue that may never have a definitive answer.  \\
\hline
\textbf{score\_task\_1 \newline content} &
2: Disagree - very little overlap of content &
2: Disagree - very little overlap of content  \\
\hline
\textbf{score\_task\_2 \newline style} &
3: Somewhat same styles &
1: Different styles  \\
\hline
\textbf{score\_task\_3 \newline reasonableness} &
5: A reasonable continuation &
5: A reasonable continuation  \\
\hline
\end{tabular}
\end{table}

\begin{table}
\caption{Annotator Scores Distribution (Examples 11-12)}
\label{tab:examples_11_12}
\centering
\small 
\begin{tabular}{|l|p{4.5cm}|p{4.5cm}|}
\hline
\textbf{} & \textbf{Example 11} & \textbf{Example 12} \\
\hline
\textbf{chat\_history} &
\textbf{user:} I'm stuck.  \newline
\textbf{assistant:} Okay. What are you stuck on?  \newline
\textbf{user:} Internships.  \newline
\textbf{assistant:} Are you struggling to find an internship?  \newline
\textbf{user:} No.  \newline
\textbf{assistant:} What's going on with internships then?  \newline
\textbf{user:} Can't decide.  &
\textbf{user:} Hey-Yo, AI!! How are you today???  \newline
\textbf{assistant:} I'm doing well, thanks. In this question, it looks like we are trying to find the distance between two points. Do you have any thoughts about how you might start something like this?  \newline
\textbf{user:} This question is sooooooo slay! Gonna slay this question. Use the distance formula.......cool, cool.  \newline
\textbf{assistant:} Great! Can you tell me what the distance formula says? How do we find the distance between two points, (x1,y1) and (x2,y2)?  \newline
\textbf{user:} Damn straight I know how to do that!! Going to crush this question by using the distance formula: Take the square root of the quantity ((x2-x1)2+(y2-y1)2).  \\
\hline
\textbf{Option A} &
Okay. I can help you decide on which internship you select.  What are you deciding between?  &
It looks like you know your distance formula - that's correct. How would you apply this to find the distance between the two points?  \\
\hline
\textbf{Option B} &
You have multiple internship options and can't decide which one to choose. That's a great problem to have! \newline \newline
What are your top options, and what are the pros and cons of each one? Maybe we can break it down together and help you make a decision.  &
YESSSS! You're absolutely on fire! That's correct! The distance formula is: \newline \newline
d = $\sqrt{((x_2 - x_1)^2 + (y_2 - y_1)^2)}$ \newline \newline
You're a math rockstar! \newline
Now, do you have a specific question you'd like to apply this formula to, or would you like me to give you a problem to solve?  \\
\hline
\textbf{score\_task\_1 \newline content} &
1: Strongly Disagree -- no content overlap  &
2: Disagree - very little overlap of content  \\
\hline
\textbf{score\_task\_2 \newline style} &
1: Different styles  &
1: Different styles  \\
\hline
\textbf{score\_task\_3 \newline reasonableness} &
5: A reasonable continuation  &
5: A reasonable continuation  \\
\hline
\end{tabular}
\end{table}

\begin{table}
\caption{Annotator Scores Distribution (Examples 13-14)}
\label{tab:examples_13_14}
\centering
\small 
\begin{tabular}{|l|p{4.5cm}|p{4.5cm}|}
\hline
\textbf{} & \textbf{Example 13} & \textbf{Example 14} \\
\hline
\textbf{chat\_history} &
\textbf{user:} I want to make an emergency kit for my house. Should I be recording this?  \newline &
\textbf{user:} I need some help.  \newline
\textbf{assistant:} Certainly. What can I help with?  \newline
\textbf{user:} Retirement.  \newline
\textbf{assistant:} Do you need help planning for retirement?  \newline
\textbf{user:} Yeah.  I'm not sure if I need to.  \\
\hline
\textbf{Option A} &
Only if you want to. I'm also happy to provide a list at the end of our conversation.  &
Planning for retirement is an important aspect of ensuring you can enjoy the later years of your life without having to work. Do you currently have anything saved for retirement?  \\
\hline
\textbf{Option B} &
which one ? &
It's never too early to start thinking about retirement planning.  Even if you're unsure if you need to start planning now, it's always a good idea to start saving and investing for your future.  Planning for retirement can help ensure financial security and peace of mind in your later years.  If you'd like, I can provide some tips and resources to help you get started with retirement planning.  \\
\hline
\textbf{score\_task\_1\_content} &
1: Strongly Disagree -- no content overlap  &
3: Neutral - some overlap of content  \\
\hline
\textbf{score\_task\_2\_style} &
1: Different styles  &
3: Somewhat same styles  \\
\hline
\textbf{score\_task\_3 \newline reasonableness} &
1: Not a reasonable continuation  &
5: A reasonable continuation \\
\hline
\end{tabular}
\end{table}

\begin{table}
\caption{Annotator Scores Distribution (Examples 15-16)}
\label{tab:examples_15_16}
\centering
\small 
\begin{tabular}{|l|p{4.5cm}|p{4.5cm}|}
\hline
\textbf{} & \textbf{Example 15} & \textbf{Example 16} \\
\hline
\textbf{chat\_history} &
\textbf{user:} Hey.  Can you help me pick a major?  \newline
\textbf{assistant:} Absolutely. Do you have a majors in mind?  \newline
\textbf{user:} Kind of.  But I'm torn. Idk what to do.  \newline
\textbf{assistant:} What majors are you considering?  \newline
\textbf{user:} My parents think I should get a degree in business, but I really want to make movies.  \newline
\textbf{assistant:} Do you want to work in entertainment?  \newline
\textbf{user:} Yeah, it sounds fun. But really hard.  \newline
  &
\textbf{user:} Do you like any shows on Netflix?  I really like Stranger Things.  \newline
\textbf{assistant:} My only purpose is to help you with your math questions.  Do you have any thoughts about how you might start this question?  \newline
\textbf{user:} What about Bridgerton, or The Queen's Gambit?  Those are both really good shows.  \newline
\textbf{assistant:} Although I'm happy that you find certain shows enjoyable, I'm here to assist you with math questions.  What do you know about solving absolute value equations?  \newline
\textbf{user:} How about Amazon Prime, or Hulu?  Do you like any shows on those?  \newline
\textbf{assistant:} My only purpose here is to assist you with the math question that I have been presented with.  Suppose I gave you an easier question, such as \textbar x\textbar  = 5. What would the solutions of that equation be?  \newline
\textbf{user:} You're no fun. But, yeah, I think I can do this.  Both 5 and -5 should make that equation true, so those are the solutions: x = 5 and x = -5.  \\
\hline
\textbf{Option A} &
A career in entertainment can be very challenging to break into. A degree in business tends to offer more flexibility in a career choice. Does that make you lean any direction?  &
That's great! In our question, we have to solve \textbar 4 - x\textbar  = 6. How could we start that question?  \\
\hline
\textbf{Option B} &
can't you just call her and ask ? &
well , that was probably because you were talking on your mobile phone . \\
\hline
\textbf{score\_task\_1\_content} &
1: Strongly Disagree -- no content overlap  &
1: Strongly Disagree -- no content overlap  \\
\hline
\textbf{score\_task\_2\_style} &
1: Different styles  &
1: Different styles  \\
\hline
\textbf{score\_task\_3 \newline reasonableness} &
1: Not a reasonable continuation  &
1: Not a reasonable continuation  \\
\hline
\end{tabular}
\end{table}

\section{LLM Usage Disclosure}
We used LLMs to check for spelling errors but did not use LLMs to directly generate any text that appear in this paper. We also used LLMs for help in minor formatting fixes of the tables.

\end{document}